\PassOptionsToPackage{dvipsnames, table, xcdraw}{xcolor}
\documentclass[twocolumn]{mic_fzu}
\usepackage{cleveref} 
\usepackage[utf8]{inputenc} 
\usepackage[T1]{fontenc} 
\usepackage[dvipsnames,table,xcdraw]{xcolor}
\usepackage{amsfonts} 
\usepackage{nicefrac}       
\usepackage{microtype}    
\usepackage{wrapfig}
 \usepackage{multirow}
 \usepackage{makecell}
 \usepackage{wasysym}
\usepackage{xcolor}
\usepackage{graphicx}
\usepackage{amsmath}
\usepackage{amssymb}
\usepackage{mathtools}
\usepackage{amsthm}
\usepackage{algorithm}
\usepackage{algorithmic}
\usepackage{makecell}
\usepackage{colortbl}
\usepackage{pifont}
\usepackage{diagbox}
\usepackage{multirow} 
\usepackage{array} 
\usepackage{graphicx, subcaption}
\usepackage{multirow}
\usepackage{colortbl}
\begin{document}
\newcommand{\nameofmethod}{OmniVaT}

\title{\nameofmethod{}: Single Domain Generalization for Multimodal Visual-Tactile Learning}

\author[1]{Liuxiang Qiu}
\author[2]{Hui Da}
\author[2]{Yuzhen Niu}
\author[1]{Tiesong Zhao$^{\dagger}$}
\author[3]{Yang Cao}
\author[3]{Zheng-Jun Zha}

\affiliation[1]{Fujian Key Laboratory for Intelligent Processing and Wireless Transmission of Media Information, College of Physics and Information Engineering, Fuzhou University, Fuzhou 350108, China} 
\affiliation[2]{College of Computer and Data Science, Fuzhou University, Fuzhou 350108, China}
\affiliation[3]{MoE Key Laboratory of Brain-inspired Intelligent Perception and Cognition, University of Science and Technology of China, Hefei 230026, China}

\affiliationenter[]{$^{\dagger}$Corresponding author.}

\abstract{
Visual-tactile learning (VTL) enables embodied agents to perceive the physical world by integrating visual (VIS) and tactile (TAC) sensors. However, VTL still suffers from modality discrepancies between VIS and TAC images, as well as domain gaps caused by non-standardized tactile sensors and inconsistent data collection procedures. We formulate these challenges as a new task, termed single domain generalization for multimodal VTL (SDG-VTL). In this paper, we propose an OmniVaT framework that, for the first time, successfully addresses this task. On the one hand, OmniVaT integrates a multimodal fractional Fourier adapter (MFFA) to map VIS and TAC embeddings into a unified embedding-frequency space, thereby effectively mitigating the modality gap without multi-domain training data or careful cross-modal fusion strategies. On the other hand, it also incorporates a discrete tree generation (DTG) module that obtains diverse and reliable multimodal fractional representations through a hierarchical tree structure, thereby enhancing its adaptivity to fluctuating domain shifts in unseen domains. Extensive experiments demonstrate the superior cross-domain generalization performance of OmniVaT on the SDG-VTL task. 
}
\mclink[Project Page]{The project page will be released.}

\maketitle
\justifying

\section{Introduction}

\begin{figure}[t]
    \centering
    \includegraphics[width=1.0 \linewidth]{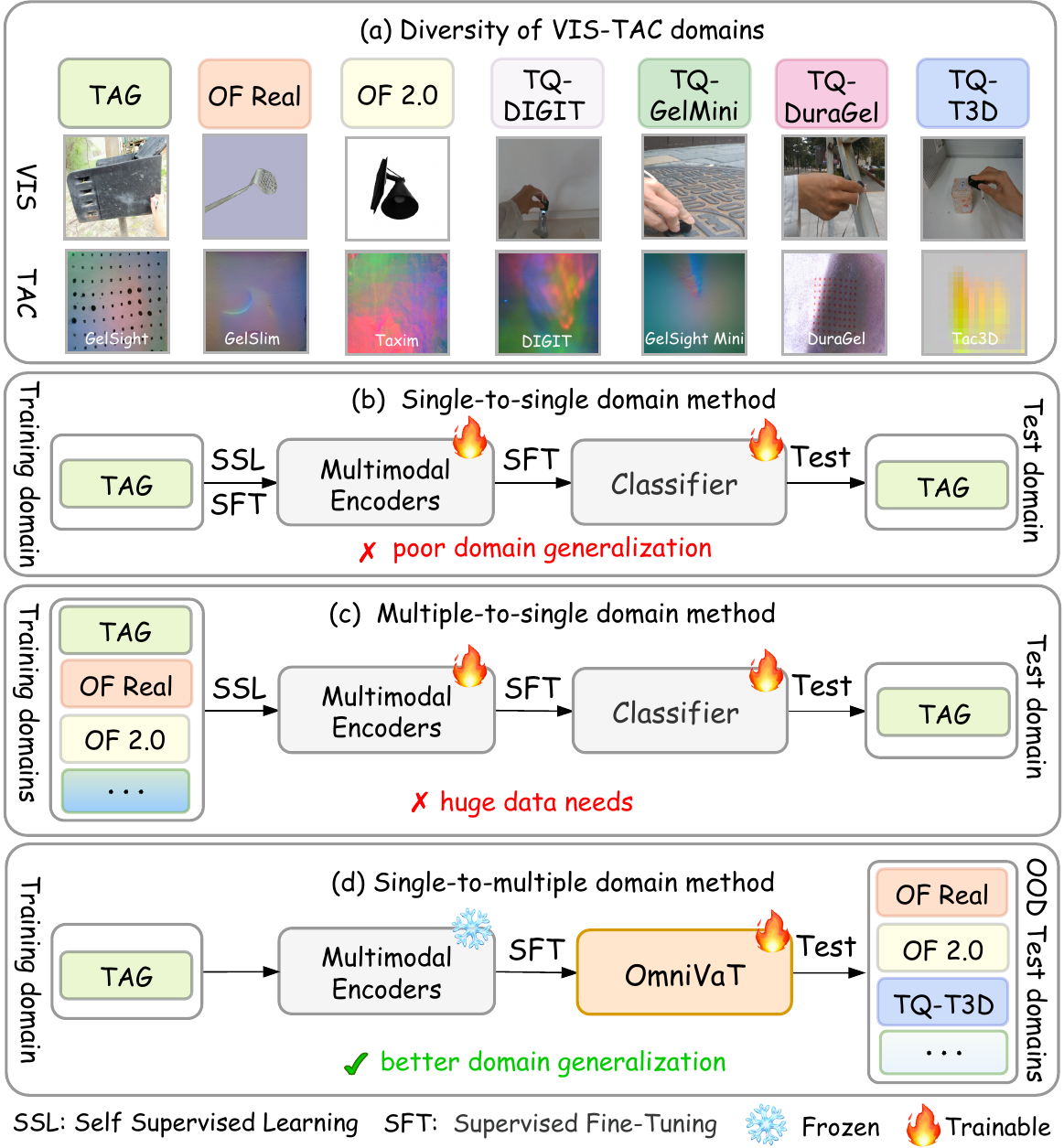}
    \caption{(a) Visualization of  various VIS-TAC domains (\eg, metal objects),  (b) Single-to-single domain methods \cite{gao2022objectfolder, yang2022touch}, (c) Multiple-to-single domain method \cite{feng2025anytouch}, (d) Our single-to-multiple domain method.}
    \label{FIG_Intro}
    \end{figure}

Recently, visual-tactile  learning (VTL), which leverages both visual (VIS) and tactile (TAC) sensors to enhance robotic perception, has been widely studied for  object understanding \cite{gao2022objectfolder, yang2024binding, feng2025anytouch, zhang2025vtla}. However, the absence of standardization in TAC sensor design leads to substantially different imaging mechanisms across sensor families, resulting in drastically varied tactile representations.  For example, the real-world GelSight sensor \cite{yuan2017gelsight} captures gel deformation under object contact using a camera assisted by LED illumination; the simulated Taxim sensor \cite{si2022taxim} employs a polynomial look-up table to approximate the optical response of the elastomer after deformation; and the Tac3D sensor \cite{zhang2022tac3d} measures the three-dimensional deformation information  using tactile skin and two cameras. Furthermore, even objects within the same category (\eg, metallic objects) exhibit diverse visual appearances due to variations in surface properties and imaging conditions, as shown in \cref{FIG_Intro}(a). These discrepancies introduce large modality and domain gaps, making it difficult to establish a unified VIS-TAC feature space. Existing VTL approaches \cite{gao2022objectfolder, yang2022touch, feng2025anytouch} are built upon the assumption that the test domain distribution is subsumed by that of the training domain, which limits their generalization ability to unseen scenarios in the real world. Here, we refer to the VIS-TAC data distribution as a specific domain which is collected by different tactile sensors under different scenarios.  For example,  single-to-single domain methods \cite{gao2022objectfolder, yang2022touch} (as given in \cref{FIG_Intro}(b)) exhibit poor generalization ability, as they are evaluated on a test domain sharing the same distribution as the training domain. Prior multiple-to-single domain method \cite{feng2025anytouch} train models on multiple VIS-TAC domains and then fine-tune them on a target domain to obtain more effective VIS-TAC representations and mitigate the modality discrepancy, as illustrated in \cref{FIG_Intro}(c).  However, these approaches require millions of training samples, making data collection and training prohibitively expensive. Motivated by the above challenges, we propose a new single domain generalization for multimodal VTL (\textbf{SDG-VTL}) task, which aims to achieve robust cross-domain VTL.

Single domain generalization (SDG) \cite{chen2023meta, liu2024stydesty, zheng2024advst, xu2025adv} aims to learn from a single source domain and generalize to multiple unseen domains, achieving impressive results in visual generalization. Though several SDG methods \cite{liu2024stydesty, zheng2024advst} employ adversarial learning to simulate cross-domain representations,  they often suffer from unstable convergence and unreliable generation. Other SDG methods  \cite{chen2023meta, gupta2025sensorinvariant} adopt metric learning to obtain domain-shared features, yet fail to effectively exploit multimodal cues, leading to suboptimal cross-modal alignment and limited generalization across unseen VIS-TAC domains. Recently, vision-language models \cite{radford2021learning, oquab2023dinov2} have shown remarkable success in improving visual generalization by learning transferable representations from large-scale text-image pairs. However, VIS sensors capture global receptive fields, while TAC sensors are inherently local and constrained to contact regions, as shown in \cref{FIG_Intro}(a). This large modality gap hinders the effectiveness of existing domain generalization methods when directly applied to cross-domain VTL.

 The existing VTL methods usually adopt cross-modal fusion \cite{wei2022alignment, li2023vito,RobustVisH2025} or common feature mapping \cite{yang2022touch, dave2024multimodal,feng2025anytouch} to mitigate the modality discrepancy. However, these methods address the discrepancy only within a single embedding space,  but they fail to learn discriminative decision boundaries for narrowing the modality gap. Several multimodal approaches \cite{zhou2025slam, chen2025bilinear, zhang2025frequency} attempt to alleviate modality discrepancies by mapping features into the frequency domain via Fourier transforms. However, insufficient exploration of the original feature space hinders the discovery of common semantic information across different modalities. Hence, the above single embedding or frequency-based methods may not be effective for achieving a discriminative multimodal  feature space, thus failing to reduce the modality gap.  Different from embedding or frequency processing, the fractional Fourier transform \cite{zhao2022fractional, yu2023deep, chen2024fractional,kocc2024trainable} maps features into a unified embedding-frequency space, providing a more effective perspective for feature projection. This facilitates the mitigation of modality and domain discrepancies among features of the same category.

To address the aforementioned challenges: \textbf{\textit{modality gap}} and \textbf{\textit{domain gap}}, we propose OmniVaT, a novel single VIS-TAC domain generalization framework, which contains a multimodal fractional Fourier adapter (MFFA) module and a discrete tree generation (DTG) module. The key novelty of our method lies in the novel formulation of \textbf{\textit{\ding{182} exploiting unified embedding-frequency representations}} and \textbf{\textit{\ding{183} expanding diverse and reliable feature nodes}} to learn a discriminative and reasonable multimodal  feature space, greatly improving  the ability of the model to mitigate modality  and  domain discrepancies. Our main contributions are summarized as follows:

\begin{itemize}

\item We observe the co-existence of the modality gap and domain gap in VTL and herein propose a new SDG-VTL task towards real-world applications.

\item  We are the first to address the SDG-VTL task with a novel OmniVaT framework that incorporates an MFFA module to align VIS-TAC features in a unified embedding-frequency space and a DTG module to discretize fractional representations into a hierarchical tree, thereby mitigating the modality  and domain gaps, respectively.

\item Our proposed OmniVaT achieves an average accuracy of 66.9\% and outperforms existing SOTA methods by at least 13.0\% across eight SDG-VTL settings.

\end{itemize}

\section{Related Work}  

\textbf{Visual-Tactile Learning (VTL).} To mitigate the modality discrepancy between VIS and TAC images, existing VTL methods can be categorized into cross-modal fusion and common feature mapping approaches. Cross-modal fusion-based methods \cite{wei2022alignment, li2023vito,RobustVisH2025} attempt to fuse VIS and TAC modality features to establish cross-modal feature associations, thus reducing the modality discrepancy. For example, Li \etal \cite{li2023vito} propose an attention-based VIS-TAC fusion network  that merges VIS and TAC features to improve recognition performance. Common feature mapping has attracted considerable attention due to its simplicity in achieving modality-shared features by metric learning \cite{yang2022touch, dave2024multimodal, feng2025anytouch}. For instance, Feng \etal \cite{feng2025anytouch} attempt to learn a multi-sensor invariant representation through self-supervised learning from huge VIS-TAC data to mitigate the modality gap. However, these methods only mitigate modality discrepancies within a single embedding space and assume identical training-test distributions, thus failing to generalize to unseen scenarios.

 \textbf{Domain Generalization.}  Domain generalization (DG) focuses on overcoming the domain shift problem, where the data distribution of the target domain differs from that of source domains used for training.  Single domain generalization (SDG) \cite{chen2023meta, zheng2024advst, liu2024stydesty, xu2025adv, gupta2025sensorinvariant} aims to achieve domain generalization when only one source domain is available. For example, Xu \etal \cite{xu2025adv} alleviate domain shift by stylization and destylization operations but suffer from unstable convergence and unreliable generation. Gupta \etal \cite{gupta2025sensorinvariant} learn domain-shared features through contrastive learning
 but fail to fully exploit multimodal cues, resulting in limited generalization to VIS-TAC domains. Several DG approaches \cite{cho2023promptstyler, zhang2025promptta, xu2025batstyler} attempt to introduce the vision-language model \cite{radford2021learning} to extract domain-shared features, yet fail to generalize to unseen TAC domains, which are absent from pre-training data. Frequency-based DG methods \cite{xu2021fourier, lu2022domain, gunduboina2025frogdognet, ji2025frequency} employ the Fourier transform for domain-invariant representation but rely on complex fusion between Fourier and embedding spaces. In contrast, the fractional Fourier transform \cite{zhao2022fractional, zhao2022cross, yu2023deep, chen2024fractional, kocc2024trainable} provides a tunable embedding-frequency projection to capture domain-invariant features. Although Zhao \etal \cite{zhao2022cross} adopt fractional fusion for spatial-spectral adaptation, their method still requires fine-tuning on target domains and lacks effective alignment to mitigate modality discrepancies in the fractional Fourier space.

\textbf{Embedding Expansion.} To enhance the model’s ability to extract discriminative domain-shared or class-shared features, a series of studies \cite{ko2020embedding, zhang2023diverse, zhang2025frequency} have attempted to generate diverse feature representations through embedding expansion to overcome the domain shift. Ko \etal \cite{ko2020embedding}  perform linear interpolation expansion between two feature points and generate synthetic points in the embedding space for metric learning. Zhang \etal \cite{zhang2025frequency} apply convolutional layers to generate diverse amplitude component nuances for visible-infrared person re-identification. The above methods expand in the embedding or frequency space, ignoring that the unified embedding-frequency space can provide a more effective class-invariant feature representation. In addition, the above methods  that rely on serial or parallel branches lack an effective mechanism to discretize embedding representations, which limits their ability to simulate various  domain shifts.

\section{Method}

\subsection{Formulation of SDG-VTL Task} Existing VTL methods \cite{yang2022touch, dave2024multimodal, feng2025anytouch} usually assume that the test domain distribution resembles the training distribution.  As a result, these methods may suffer from overfitting to seen domain features and fail to effectively recognize objects from unseen VIS-TAC domains due to the diversity of the  VIS-TAC domains. Inspired by the widely recognized single domain generalization \cite{liu2024stydesty, zheng2024advst,  xu2025adv}, we introduce the  single domain generalization for multimodal VTL (SDG-VTL) task  that minimizes the single source VIS-TAC domain $\mathcal{S}$ risk while controlling the worst-case risk over unseen target domains $\mathcal{T}$, that is
\begin{equation}
 \min_{\theta} \;
\sup_{{\mathcal{T}}:{G}({\mathcal{S}}, \mathcal{T})\leq \rho }
\; \mathbb{E}_{(x^{v}, x^{t}, y) \sim \mathcal{T}}
\, [\ell ({\theta}; (x^{v}, x^{t}, y))],
\end{equation}
where   $G$ is a similarity metric to measure the domain distance and $\rho$ denotes the largest domain discrepancy between $\mathcal{S}$ and $\mathcal{T}$;  $x^{v}$ and $x^{t}$ are VIS  and TAC features, respectively; $y$ is the ground-truth label; $\theta$ is the parameters of the network; and $\ell(;)$ is the task evaluation (\eg, cross-entropy loss). 

   \begin{figure}[t]
    \centering
    \includegraphics[width=1.0 \linewidth]{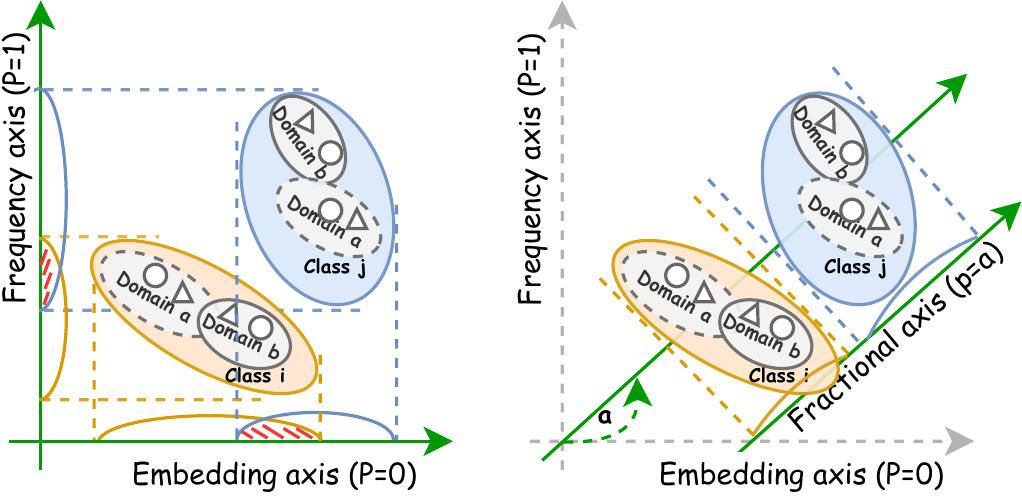}
    \caption{Embedding ($p=0$) or frequency ($p=1$) processing causes overlapping projections and inter-class confusion.  However, the fractional Fourier transform (FrFT, $p=a$) provides a unified embedding-frequency representation that separates features from different classes more effectively. Circles and triangles represent VIS and TAC features, respectively.}
    \label{FRFT}
    \end{figure}

    \begin{figure*}[t] 
   \centering
    \includegraphics[width=0.9\linewidth]{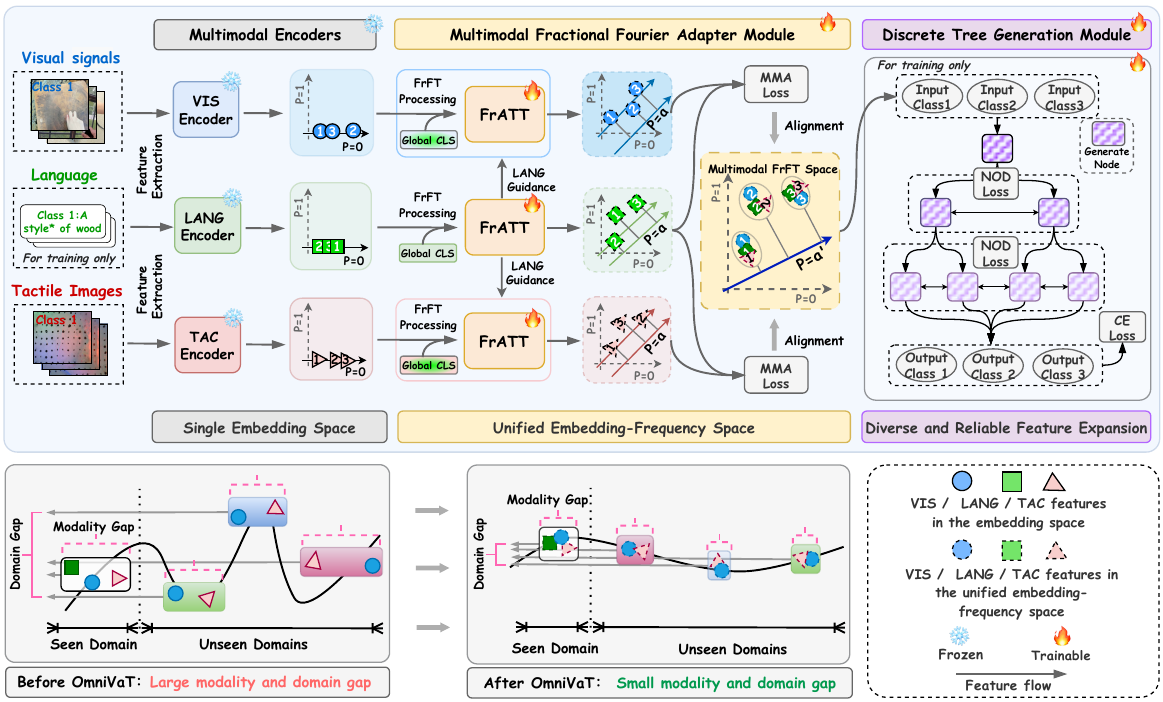} 
    \caption{Overview of the proposed OmniVaT, including frozen multimodal encoders, a multimodal fractional Fourier adapter (MFFA) module with the fractional Fourier attention (FrATT) mechanism, and a discrete tree generation (DTG) module. The OmniVaT is jointly optimized by a multimodal alignment (MMA) loss, a node diversity (NOD) loss, and a cross-entropy (CE) loss. The frozen encoders are derived from CLIP \cite{radford2021learning}. Both the MFFA and DTG modules share the same set of parameters across modalities. Since the LANG embedding contains class information, we use it only during training.}
    \label{fig:overview}
\end{figure*}

\subsection{Motivation} As shown in \cref{FIG_Intro}(a),  the modality and domain gaps are the main issues that hinder existing domain generalization and VTL methods from effectively learning discriminative features.  The fractional Fourier transform (FrFT) \cite{zhao2022fractional, yu2023deep, chen2024fractional,kocc2024trainable} allows flexible transformations between embedding and frequency spaces at arbitrary angles, providing a more powerful mechanism for capturing the semantic consistency of representations, as shown in \cref{FRFT}. This can be considered an effective approach to achieve class-irrelevant features. The $p$-th order FrFT  of the embedding $\mathbf{E} \in \mathbb{R}^{1 \times D}$, is defined as follows:
\begin{equation}
{\rm FrFT}_{p}(\mathbf{E}(u_0)) = \int_{-\infty}^{\infty} K_p(u_0, u_p) \mathbf{E}(u_0) \, du_0,
\end{equation}
where $K_p(u_0, u_p)$ is the kernel function of the fractional Fourier transform; $u_0$ refers to the original embedding space, and  $u_p$ represents the transformed fractional space.  \textbf{More details about FrFT can be seen in the supplements.}

We then design modules to address modality and domain gaps based on FrFT.  For  \textbf{modality gap:}  existing VTL approaches \cite{yang2022touch, RobustVisH2025}, which mainly rely on feature fusion or metric alignment, often suffer from careful module design and insufficient semantic exploitation. Language (LANG) information provides high-level abstract representations of categories that can help the network achieve modality-shared features. Thus, we aim at designing a multimodal fractional Fourier adapter (MFFA) module, which incorporates FrFT to map the VIS-TAC feature into the unified fractional space with the guidance of the  LANG information to mitigate the modality discrepancy.  LANG information is used only during training because it contains class information and therefore cannot be applied during inference.

For \textbf{domain gap:}  existing  single domain generalization methods typically leverage adversarial learning or feature augmentation to increase the domain diversity. However, due to the instability of the stylization and destylization process, this will result in the inability to generate reliable features. Though several studies \cite{ko2020embedding,zhang2023diverse,zhang2025frequency} expand embeddings to mitigate domain shift, they follow either serial or parallel expansion designs and struggle to effectively generate diverse feature representations. The tree-based learning methods \cite{yao2023tree, miao2024specinfer, yang2025discrete} ensure the continuity of extended features while maintaining the diversity among  features, and naturally emerge as an effective solution for learning various feature representations, thereby improving the generalization performance of the model. Based on the above observations, we propose a discrete tree generation (DTG) module to explore and learn a diverse fractional representation set, thereby enhancing the model’s single domain generalization ability to unseen domains.

\subsection{Overview of Proposed OmniVaT}
The overview of our proposed OmniVaT  is shown in  \cref{fig:overview}. OmniVaT consists of  frozen VIS, TAC, and LANG encoders, an MFFA module, and a DTG module, respectively. We adopt PromptStyler \cite{cho2023promptstyler} to generate the missing language information (\eg, “A style* of [Class]”).

\subsection{Multimodal Fractional Fourier Adapter (MFFA) Module}
Suppose that we have the single source VIS-TAC domain ${\mathcal{D}_{s}}=\{x_{i}^{v}, x_{i}^{t},  y_{i}\}_{i=1}^{N}$ with  auxiliary LANG  information ${x_{i}}^{l}$,  where $N$ represents number of samples. After passing through the frozen VIS encoder, TAC encoder, and LANG encoder, we obtain the  VIS embedding $\mathbf{E}^{v} \in \mathbb{R}^{1 \times D}$,  TAC embedding $\mathbf{E}^{t} \in \mathbb{R}^{1 \times D}$,  and LANG embedding $\mathbf{E}^{l} \in \mathbb{R}^{1 \times D}$, respectively. Here, $D$  is the dimension of the embedding.  To reduce the modality gap, we propose the MFFA module, which maps VIS and TAC embeddings into a fractional embedding-frequency space with the guidance of the LANG information by the cross-modal fractional Fourier attention mechanism, and further mitigates modality discrepancies through the multimodal alignment loss. This module eliminates the need for careful cross-modal fusion design while enabling more comprehensive information mining across modalities. Firstly, we introduce the FrFT processing to map the LANG embedding into the unified fractional space, that is
\begin{equation}
\begin{aligned}
\mathbf{F}^{l}&={\rm FrFT~ Processing}(\mathbf{E}^{l})\\&={{\mathcal{U}}}({\mathcal{R}}({\rm FrFT}_{p}(\theta_{e,l} {{\mathbf{E}}}^{l})))+{j}~{{\mathcal{U}}}({{\mathcal{I}}}({\rm FrFT}_{p}(\theta_{e,l} {{\mathbf{E}}}^{l}))),
\end{aligned}
\label{fr}
\end{equation}
where  $\theta_{e,l}\in \mathbb{R}^{E \times 1}$ are trainable parameters that expand the embedding dimensionality,  ${\rm FrFT}_{p}(\cdot)$ is the $p$-th order fractional Fourier transform operation; $\mathcal{R}(\cdot)$ and $\mathcal{I}(\cdot)$ denote the real and imaginary part operators, respectively; $\mathcal{U}(\cdot)$  means the ReLU operation; $j$ is the imaginary unit. 

Then, we propose a fractional  Fourier attention (FrATT) mechanism to achieve  the encoded fractional LANG representation, which can be written as:
\begin{equation}
\begin{aligned}
& \mathbf{\bar{F}}^{l} = {\rm FrATT}(\mathbf{{F}}^{l}, \mathbf{F}^{l}_{g}\oplus\mathbf{F}^{l})
\\ &={\rm FFN}({{\rm Mean}}(\frac{{{\rm SM}}(\theta_{Q}{\mathbf{F}}^{l} \theta_{K} {(\mathbf{F}^{l}_{g}\oplus\mathbf{F}^{l})}^{\rm T})}{\sqrt{D}}\theta_{V}({\mathbf{F}^{l}_{g}\oplus \mathbf{F}^{l}}))),
\end{aligned}
\end{equation}
where global class (CLS) token $\mathbf{F}^{l}_{g} \in \mathbb{R}^{1 \times D}$ means the average representation of the same class of the input $\mathbf{F}^{l}$ in the mini-batch which can get the global information of class to overcome intra-class variations; $\oplus$ denotes the broadcast addition operation; $\theta_{Q} \in \mathbb{R}^{D \times D}$, $\theta_{K} \in \mathbb{R}^{D \times D}$, and $\theta_{V} \in \mathbb{R}^{D \times D}$ are linear layers to project the $\mathbf{F}^{l}$ and $(\mathbf{F}^{l}_{g} \oplus\mathbf{F}^{l})$  into query space, key, and value space, respectively; ${\rm SM}(\cdot)$ and ${\rm Mean}(\cdot)$ denote the Softmax and mean operations, respectively. ${\rm FFN(\cdot)}$ is the fully connected feedforward network. Subsequently, guided by the LANG feature $\mathbf{\bar{F}}^{l}$, we further project the VIS embedding $\mathbf{E}^{v}$ into the unified  embedding-frequency feature space to achieve modality-invariant representations, that is
\begin{equation}
\begin{aligned}
\mathbf{F}^{v}&={\rm FrFT} {~\rm Processing}(\theta_{e,{v}} ({\rm {\mathbf{\bar{F}}}}^{l}+\mathbf{E}^{v})),\\
\end{aligned}
\label{frp}
\end{equation}
where  $\theta_{e,{v}} \in \mathbb{R}^{E \times 1}$ is the trainable parameters that extend the number of embedding. We further adopt the FrATT mechanism to establish semantic correlations among different modalities, as follows
\begin{equation}
\begin{aligned}
\mathbf{\bar{F}}^{v} &= {\rm FrATT}(\mathbf{\bar{F}}^{l}, \mathbf{F}^{v}_{g}\oplus \mathbf{F}^{v}).
\end{aligned}
\label{fratt}
\end{equation}
Similarly to \cref{frp,fratt}, we can also obtain the fractional TAC feature $\mathbf{\bar{F}}^{t}$ after language guidance and the FrATT mechanism. To better optimize the fractional order in the  embedding-frequency space, we align the representations between VIS, TAC, and LANG features to reduce the modality gap by the multimodal alignment (MMA) loss
\begin{equation}
\begin{aligned}
\mathcal{L}_{{\rm MMA}} =  \lambda (D_{{\rm KL}}(\mathbf{\bar{F}}^{l}, \mathbf{\bar{F}}^{v})  + D_{{\rm KL}}(\mathbf{\bar{F}}^{l}, \mathbf{\bar{F}}^{t})),
\end{aligned}
\end{equation}
where $D_{{\rm KL}}(\cdot)$ denotes the Kullback-Leibler (KL) divergence function, $\lambda$ adjusts the contribution of KL divergence.  After the MFFA module, we can efficiently and effectively exploit the associations among different modality embeddings in the unified embedding-frequency space, thus greatly mitigating the modality discrepancy.

\subsection{Discrete Tree Generation (DTG) Module}
To reduce the domain shift, we propose a DTG module, which hierarchically expands diverse and reliable features in the unified embedding-frequency space with a tree structure. Specifically, given an input unified embedding-frequency VIS-TAC-LANG feature $\mathbf{{T}}$ (\ie, $\mathbf{\bar{F}}^{v}+\mathbf{\bar{F}}^{t}+\mathbf{\bar{F}}^{l}$), we construct a binary tree where each node corresponds to an expanded feature. The child nodes are generated through a linear transformation function $\mathcal{F}(\cdot)$ applied to the parent node, allowing multi-level expansion. Formally, the expansion process at layer $r$ can be written as  
\begin{equation}
   {\mathbf{T}_{m,n}^{(r+1)}} = \mathcal{F}({\mathbf{{T}}_{m}^{(r)}}, \mathbf{W}^{(r)}_{m,n}),
\end{equation}
where ${\mathbf{{T}}_{m}^{(r)}}$ denotes the $m$-th parent feature node at layer $r= \{1, \cdots, R-1\}$ and $R$ is the depth of the tree;  $\mathbf{W}^{(r)}_{m,n}  \in \mathbb{R}^{D \times D}$ are learnable parameters  for the $n$-th child node of $m$-th parent  node.  By recursively applying $\mathcal{F}(\cdot)$, we obtain a expanded fractional feature set of each layer $\mathbf{T}^{tree} = \{{\mathbf{T}}^{(r)}_{m}\}$.  

\begin{table*}[t]
  \centering
  \caption{Comparison with state-of-the-art methods on the SDG-VTL across eight VIS-TAC domains. “$\mathcal{S}$" / “$\mathcal{T}$" means the source / target domain, and “$\rightarrow$ X”  represents the evaluation of the other unseen target domains, excluding the seen source domain. ${\lozenge}$   and ${\Join}$ denote original TAC learning and VTL methods, respectively. The best results are marked with \textbf{Bold}.  \textcolor{red}{$\Delta$ LDC}  denotes the improvements of the proposed OmniVaT over  LDC. The metrics are accuracy (ACC, \%) and Macro-F1 (F1*, \%), respectively.  }
   \resizebox{1.0\textwidth}{!}{
  \begin{tabular}{rccccccccc}
  \hline
  
   \hline
  \multirow{2}{*}{Methods  / {\small Venue}} & {\makecell[c]{$\mathcal{S}$: TAG\\ $\mathcal{T}$: $\rightarrow$ X}} &{\makecell[c]{$\mathcal{S}$: OF Real\\ $\mathcal{T}$: $\rightarrow$ X}} & {\makecell[c]{$\mathcal{S}$: OF 2.0 A\\ $\mathcal{T}$: $\rightarrow$ X}} & {\makecell[c]{$\mathcal{S}$: OF 2.0 B \\$\mathcal{T}$: $\rightarrow$ X}}&{\makecell[c]{$\mathcal{S}$: TQ-DIGIT\\$\mathcal{T}$: $\rightarrow$ X}}&{\makecell[c]{$\mathcal{S}$: TQ-GelMini\\$\mathcal{T}$: $\rightarrow$ X}} & {\makecell[c]{$\mathcal{S}$: TQ-DuraGel\\ $\mathcal{T}$: $\rightarrow$ X}}&{\makecell[c]{$\mathcal{S}$: TQ-T3D \\$\mathcal{T}$:  $\rightarrow$ X}} &{Average}\\ 
  \cline{2-10}
  ~ &  ACC  / F1* & ACC / F1* & ACC / F1* & ACC / F1*& ACC / F1*& ACC / F1*& ACC / F1* & ACC / F1*& ACC / F1* \\ 
        \hline
            \hline
 SITR$^{\lozenge}$ /  {\small ICLR-25} \cite{gupta2025sensorinvariant}& 33.1 / 16.7&25.3 / 14.7&34.7 / 19.8 &20.1 / 9.7& 30.6 / 16.8&33.6 / 33.6 &34.1 / 22.3&38.4 / 25.5& 31.2 / 19.9\\

VT CMC$^{\Join}$  /  {\small NeurIPS-22}  \cite{yang2022touch}&  31.5 /  10.7 & 25.8 / 9.8& 33.9 / 6.3& 32.9 / 5.7& 36.0 / 21.4& 28.5 / 13.1& 31.7 / 19.2& 15.5 / 7.3  &29.5 / 11.7\\
  RobustVisH$^{\Join}$  /   {\small MM-25}  \cite{RobustVisH2025} & 32.7 / 17.9& 30.9 / 10.8& 32.9 / 11.6& 32.5 / 11.5& 27.1 / 10.4& 27.5 / 10.7& 28.1 / 11.9& 23.5 / 9.7 &29.4 / 11.8\\

    \hline
    \hline

    ~ &    \multicolumn{9}{c}{\textit{ResNet-50 \cite{he2016deep} with pre-trained weights from CLIP \cite{radford2021learning}}} \\
            \hline
CLIP /  {\small ICML-21}   \cite{radford2021learning} &40.7 / 33.4&41.4 / 34.8&40.6 / 34.5 &40.6  / 34.5& 39.1 / 30.2& 39.8 / 30.2 & 38.7 / 30.2 & 29.6 / 22.5 & 38.8 / 31.3\\

  PromptStyler  /  {\small ICCV-23} \cite{cho2023promptstyler} &48.7 / 37.3 &36.3 / 25.3& 45.5 / 38.7& 34.9 / 29.8& 44.4 / 35.5& 46.1 / 38.4& 45.0 / 39.3&  47.0 / 31.8 &43.5 / 34.5\\
  SPG /  {\small ECCV-24} \cite{bai2024soft} &47.9 / 43.3&42.8 / 38.7&45.2 / 42.9&38.5 / 29.7&58.7 / 52.5 & 56.4 / \textbf{54.7} & 55.2 / 51.0 & 47.1 / 35.6  & 49.0 / 43.6\\
    LDC /  {\small CVPR-25} \cite{li2025logits}&43.9 / 33.3&43.2 / 32.7&44.8 / 33.8&52.3 / 43.7 & 38.3 / 22.0& 37.1 / 22.6& 39.4 / 25.6& 43.9 / 27.9 &42.9 / 30.2\\
            \hline

  OmniVaT / {\small Ours}&\textbf{52.0} / \textbf{49.7}&\textbf{{51.6}} / \textbf{50.0}&\textbf{52.4} / \textbf{52.3}& \textbf{53.0} / \textbf{55.2}& \textbf{59.1} / \textbf{54.6}& \textbf{57.6} / {52.7}& \textbf{58.4} / \textbf{52.9}& \textbf{52.4} / \textbf{39.8}  & \textbf{54.6} / \textbf{50.9} \\   

\textcolor{red}{$\Delta$ LDC}& \textcolor{red}{8.1} / \textcolor{red}{16.4}& \textcolor{red}{8.4} / \textcolor{red}{17.3}& \textcolor{red}{7.6} / \textcolor{red}{18.5}&\textcolor{red}{0.7} / \textcolor{red}{11.5}&\textcolor{red}{20.8} / \textcolor{red}{32.6}&\textcolor{red}{20.5} / \textcolor{red}{30.1}&\textcolor{red}{19.0} / \textcolor{red}{27.3}&\textcolor{red}{8.5} / \textcolor{red}{11.9}&\textcolor{red}{11.7} / \textcolor{red}{20.7} \\ 
     	 	 	 
      \hline
      \hline
   ~ &    \multicolumn{9}{c}{\textit{ViT-B / 16 \cite{dosovitskiy2020image} with pre-trained weights  from CLIP \cite{radford2021learning}}} \\
    \hline
CLIP /  {\small ICML21} \cite{radford2021learning}& 47.8 / 40.0&48.2 / 40.1& 47.8 /  41.1& 47.8 / 41.1& 48.3 / 36.7 & 46.5 / 37.4 & 46.5 / 35.0 & 42.6 / 30.8& 46.9 / 37.8 \\

  PromptStyler / {\small ICCV-23} \cite{cho2023promptstyler}&51.7 / 40.6 &42.3 / 30.1& 48.9 / 41.2&38.0 / 26.7& 54.3 / 47.2& 47.0 / 39.8& 51.8 / 48.1& 49.4 /  30.9 & 47.9 / 38.1\\

    SPG  /  {\small ECCV-24} \cite{bai2024soft}&49.4 / 41.9&49.8 / 42.0&49.1 / 42.8&49.4 / 42.0 & 53.1 / 46.9 & 53.8 / 48.6 &51.6 / 46.3 & 50.4 / 43.8& 50.8 / 44.3\\

    MMRL /  {\small CVPR-25} \cite{guo2025mmrl}&46.0 / 33.5 & 45.9 / 45.4 &42.9 / 30.1 &42.3 / 37.2 &36.0 / 17.9 & 27.3 / 15.2 & 29.6 / 14.4& 48.8 / 34.8 & 39.9 / 28.6\\
    LDC /  {\small CVPR-25} \cite{li2025logits}&50.1 / 36.3&49.5 / 37.6&52.2 / 43.9&52.2 / 43.8& 53.8 / 38.9& 44.7 / 35.3& 49.2 / 38.3& 52.9 / 34.9 &50.6 / 38.6 \\
            \hline

    OmniVaT / {\small Ours}&\textbf{{56.2}} / \textbf{53.7}&\textbf{58.4} / \textbf{56.5}&\textbf{58.6} / \textbf{58.2}&\textbf{61.2} / \textbf{60.3}&\textbf{56.4} / \textbf{50.1}&\textbf{57.3} / \textbf{53.5}& \textbf{56.3} / \textbf{51.8} & \textbf{54.1} / \textbf{47.9} &\textbf{57.3} / \textbf{54.0}\\

\textcolor{red}{$\Delta$ LDC}& \textcolor{red}{6.1} / \textcolor{red}{17.4}&\textcolor{red}{8.9} / \textcolor{red}{18.9}& \textcolor{red}{6.4} / \textcolor{red}{14.3}& \textcolor{red}{9.0} / \textcolor{red}{16.5}& \textcolor{red}{2.6} / \textcolor{red}{11.2} &\textcolor{red}{12.6} / \textcolor{red}{18.2} & \textcolor{red}{7.1} / \textcolor{red}{13.5} &\textcolor{red}{1.2} / \textcolor{red}{13.0}&\textcolor{red}{6.7} / \textcolor{red}{15.4}\\ 

      \hline
            \hline
   	 	 	 
   ~ &    \multicolumn{9}{c}{ \textit{ViT-L / 14 \cite{dosovitskiy2020image} with pre-trained weights  from CLIP \cite{radford2021learning}}} \\
            \hline
  CLIP  /  {\small ICML-21} \cite{radford2021learning} & 48.7 / 42.6&48.9 / 42.8&49.8 / 44.8 & 49.8 / 44.8&  48.9 / 39.3 &  46.7 / 38.3&  46.1 / 37.9&  39.7 / 30.7 &47.3 / 40.1\\
  PromptStyler  /  {\small ICCV-23} \cite{cho2023promptstyler} &54.4 / 47.6 & 42.6 / 34.9& 49.8 / 47.1& 44.7 / 35.6& 55.5 / 46.7& 54.3 / 46.4& 54.6 /  49.4& 49.8 / 33.3 &50.7 / 42.6\\
        SPG  /  {\small ECCV-24} \cite{bai2024soft} &57.5 / 54.5&53.6 / 49.7&47.8 / 39.8&39.0 / 31.9 &61.0 / 53.4&55.7 / 49.3&65.0 / 61.8& 51.9 / 46.3 &53.9 / 48.3 \\
    MMRL /  {\small CVPR-25} \cite{guo2025mmrl}&59.5 / 54.6 &35.2 / 35.1 &34.3 / 22.9 &44.6 / 39.4 & 40.6 / 41.4 & 33.3 / 22.8 & 30.1 / 19.6 & 51.2 / 43.2 &41.1 / 34.9\\
    LDC /  {\small CVPR-25} \cite{li2025logits}&55.2 / 45.5&49.2 / 43.2&52.7 / 43.5&55.3 / 46.7 &53.9 / 38.9 &53.6 / 42.0 & 55.3 / 41.6 &53.5 / 35.2 &53.6 / 42.1\\
 \hline
OmniVaT / {\small Ours}&\textbf{62.6} / \textbf{64.1}&\textbf{{64.9}} / \textbf{65.8} &\textbf{67.4} / \textbf{68.0}&  \textbf{67.2} / \textbf{68.5} &\textbf{68.5} / \textbf{65.9}&\textbf{68.8} / \textbf{67.2}&\textbf{70.1} / \textbf{66.8}& \textbf{65.4} / \textbf{58.0} &\textbf{66.9} / \textbf{65.5}\\ 

\textcolor{red}{$\Delta$ LDC} &\textcolor{red}{7.4} / \textcolor{red}{18.6}	&\textcolor{red}{15.7} / \textcolor{red}{22.6}	&\textcolor{red}{14.7} / \textcolor{red}{24.5}	&\textcolor{red}{11.9 / 21.8}	&\textcolor{red}{14.6 / 27.0}	&\textcolor{red}{15.2} / \textcolor{red}{25.2}	&\textcolor{red}{14.8} /  \textcolor{red}{25.2}	&\textcolor{red}{11.9}  / \textcolor{red}{22.8} & \textcolor{red}{13.3} / \textcolor{red}{23.4}\\ 

      \hline
             
             \hline
  \end{tabular}
  
}
        \label{T-Comparison}

\end{table*}

To further enhance feature diversity and prevent redundancy among expanded nodes, we introduce node diversity (NOD) loss  as a regularization term.  By minimizing  NOD loss (\ie, reducing the cosine similarity between different node representations),   we encourage their independence and thus achieve a more dispersed exploration of the feature space. Concretely, the cosine similarity ${\mathbf{A}}^{(r)}_{i,j}$ between $i$-th  node  $\mathbf{T}_{i}^{(r)}$ and $j$-th node $\mathbf{T}_{j}^{(r)}$ at layer $r$ can be computed as  
\begin{equation}
    {\mathbf{A}}^{(r)}_{i,j} = \frac{{\mathbf{{T}}_{i}^{(r)}}^\top  {\mathbf{{T}}_{j}^{(r)}}}{\ell_2({\mathbf{{T}}_{i}^{(r)}}) \ell_2({\mathbf{{T}}_{j}^{(r)}})},
\end{equation}
where $\ell_{2}(\cdot)$ means  L2-normalization operation.
We incorporate the NOD loss to hierarchically discretize node representation at different layers, as follows

\begin{equation}
    \mathcal{L}_{{\rm NOD}} = \sum\nolimits_{r=1}^{R} \|\mathbf{A}^{(r)} - \mathbf{I}^{(r)}\|_{F}.
\end{equation}
where  $\mathbf{I}^{(r)}$ is the identity matrix  and whose shape is $2^{{r-1}}\times 2^{{r-1}}$; $\|\cdot\|_F$ denotes the Frobenius norm  that provides stable gradients and facilitates more reliable optimization. The NOD loss promotes a structured and progressive disentanglement of node features, and the domain representation enhanced VIS  and TAC  features, \ie $\mathbf{\hat{F}}^{v}$ and $\mathbf{\hat{F}}^{t}$, that is
\begin{equation}
 \mathbf{\hat{F}}^{v} = {{\rm Mean}}(\mathbf{T}^{tree}) + \mathbf{\bar{F}}^{v},  \mathbf{\hat{F}}^{t} = {{\rm Mean}}(\mathbf{T}^{tree}) + \mathbf{\bar{F}}^{t}.
\end{equation}

In this way, the proposed DTG module enables a structured and progressive exploration of diverse feature spaces, producing robust modality-invariant representations while avoiding the oversimplified parallel expansion of previous embedding-based approaches.   The joint loss is defined as
\begin{equation}
    \mathcal{L} = \mathcal{L}_{{\rm MMA}} + \mathcal{L}_{{\rm NOD}} + \mathcal{L}_{{\rm CE}},
\end{equation}
where $\mathcal{L}_{{\rm CE}}$ represents the cross-entropy loss.

\section{Experiments}

\textbf{More details on datasets, task settings, ablation studies, and comparison results are provided in the supplements.}

\subsection{Experimental Settings}
\textit{\textbf{1) Datasets and evaluation metrics:}}
We collect VIS-TAC images from four datasets (\ie, Touch-and-Go (TAG) \cite{yang2022touch}, ObjectFolder (OF) Real \cite{gao2023objectfolder}, ObjectFolder (OF) 2.0 \cite{gao2022objectfolder}, and TacQuad (TQ) \cite{feng2025anytouch}) spanning 8 distinct domains (\ie, TAG, OF Real, OF 2.0 A, OF 2.0 B, TQ-DIGIT, TQ-GelMini, TQ-DuraGel, and TQ-T3D), that share five common material categories (\ie, wood, steel, plastic, ceramic, and glass).  We train on a single source domain and evaluate on the remaining unseen target domains, reporting the average top-1 accuracy (ACC) and Macro-F1 (F1*) over three random seeds.

\textit{\textbf{2) Implementation Details:}} During the training step, VIS and TAC images are resized to 3 $\times$ 224 $\times$ 224 and augmented by the random horizontal flip and the feature normalization. We choose the publicly available pre-trained model CLIP \cite{radford2021learning} as our frozen  VIS,  TAC, and LANG encoders.  VIS  and the TAC encoders are implemented with ResNet-50 \cite{he2016deep}, ViT-B/16  \cite{dosovitskiy2020image}, or ViT-L/14  \cite{dosovitskiy2020image}, the LANG encoder is adopted by Transformer in experiments. The dimensions of each LANG, VIS, and TAC embedding are D = 1024 / 512 / 768 when VIS and the TAC encoders  are implemented with ResNet-50  / ViT-B/16 / ViT-L/14, respectively. The fractional order $p$, extension parameter $E$,  balance weight $\lambda$, and depth $R$ of the tree  are set to 0.5, 4, 10, and 3 in the FrFT and DTG modules, respectively. As PromptStyler \cite{cho2023promptstyler}, we generate the 80 LANG samples  for each class. For each mini-batch, 16 paired VIS-TAC samples with auxiliary LANG information  are randomly selected. We introduce the cosine annealing  warm-up strategy \cite{loshchilov2017sgdr} to update the learning rate, and the initial learning rate is set to 0.05. OmniVaT is optimized by stochastic gradient descent with a momentum of 0.9 and trained for 20 epochs, implemented in PyTorch on an NVIDIA RTX 3090 GPU.
       \begin{table}[!t]
                  \caption{Cosine similarity margins of different multimodal learning methods which are trained on TAG and tested on the unseen OF Real, OF 2.0 A, and TQ-T3D domains.}
  \begin{center}
    \resizebox{0.47\textwidth}{!}{
    \begin{tabular}{c | c | c | c   | c | c   }
    \hline   
    \multirow{1}*{Methods}& OF Real & OF 2.0 A&TQ-T3D &\makecell{[OF Real; \\ TQ-T3D]}&\makecell{[OF 2.0 A; \\ TQ-T3D]}\\

    \hline  
        CLIP  \cite{radford2021learning}&0.036&0.075 &0.033&0.007 &0.006\\
        PromptStyler \cite{cho2023promptstyler}&0.034&0.061 &0.035&0.013 &0.017\\
        SPG \cite{bai2024soft}&0.029&0.050 &0.026&0.016 &0.020\\
       MMRL \cite{guo2025mmrl}&0.016&0.030 &0.022&0.008 &0.015\\
       LDC  \cite{li2025logits}&0.041&0.040&0.046& 0.028 &0.017\\
       OmniVaT &0.168&0.258&0.218& 0.191&0.143\\
    \hline 
    \end{tabular}
    }
        \label{T-SM}
    \end{center}
    \end{table}

   \begin{table*}[!t]
                  \caption{All methods are trained and tested on a single NVIDIA RTX 3090 GPU using the TAG domain.  Trainable parameters (M),  training time (s/epoch) on the TAG dataset, GPU memory (MB), inference speed (FPS),   accuracy (ACC, \%), and Macro-F1 (F1*, \%) are reported. TAG $\rightarrow$ X means that the methods are trained on the TAG domain and tested on the other unseen domains.  $^{\ddag}$ means that we reduce the feature dimension to 64.   CLIP \cite{radford2021learning} is the zero-shot method.}
  \begin{center}

    \resizebox{0.9\textwidth}{!}{
    \begin{tabular}{r c  c  c   c c c}
    \hline   
    \multirow{1}*{Methods  / Venue} & \makecell{Trainable\\ Parameters}&\makecell{Training\\Time}& \makecell{GPU\\Memory}&\makecell{Inference\\Speed}&\makecell{TAG $\rightarrow$ X\\ACC / F1*}&\makecell{OF 2.0 A $\rightarrow$ X\\ACC / F1*} \\
    \hline  
       ~ &     \multicolumn{6}{c}{\textit{ViT-B / 16 \cite{dosovitskiy2020image} with pre-trained weights  from CLIP \cite{radford2021learning}}} \\
    \hline  

       CLIP  /  {\small ICML-21}  \cite{radford2021learning}&-  &- &1,158&188.2&47.8 / 40.0&47.8 / 41.1\\
       PromptStyler /  {\small ICCV-23} \cite{cho2023promptstyler}&0.3   &34.1&2,555&156.8&51.7 / 40.6&48.9 / 41.2\\
       SPG  /  {\small ECCV-24} \cite{bai2024soft}&4.7&23,180.2&4,936&91.1&49.4 / 41.9&49.1 / 42.8\\
	MMRL /  {\small CVPR-25} \cite{guo2025mmrl}&5.0 &58.3 &3,603&163.3&46.0 / 33.5&42.9 / 30.1\\
       LDC  /  {\small CVPR-25} \cite{li2025logits}&3.4 &23.2&2,011&172.5&50.1 / 36.3&52.2 / 43.9\\
        OmniVaT /  {\small Ours}&5.1&141.9&2,845&75.2&56.2 / 53.7 &58.6 / 58.2\\
        OmniVaT $^{\ddag}$ /  {\small Ours}&0.2&98.4&2,779&111.1&55.0 / 54.1&56.3 / 55.1\\

    \hline 
    \end{tabular}
    }
        \label{T-CC}
    \end{center}
    \end{table*}

\subsection{Comparison with State-of-the-Art Methods}
 \textit{\textbf{1) Effectiveness of SDG-VTL:}}
\cref{T-Comparison} shows that, compared with previous common feature mapping  VTL method (\ie,  VT CMC \cite{yang2022touch}), the  proposed OmniVaT improves the recognition accuracy  by 20.5\%  on the TAG $\rightarrow$ X task. 
 OmniVaT also achieves the  Macro-F1 of 50.0\%, about 39.0\% higher than  RobustVisH \cite{RobustVisH2025} which introduces cross-modal fusion to integrate multimodal features on the OF Real $\rightarrow$ X task. Although SITR \cite{gupta2025sensorinvariant} aims to learn sensor-invariant representations through pre-training on large-scale simulated TAC data, our proposed OmniVaT surpasses SITR  at least by 17.7\% in accuracy when trained on the simulated to real-world scenarios (\eg,  OF 2.0 A $\rightarrow$ X).  \cref{T-Comparison}  also shows that our OmniVaT outperforms the previous multimodal learning methods MMRL  \cite{guo2025mmrl} and  LDC  \cite{li2025logits} by about 13.0\% Macro-F1 with pre-trained ViT-B/16 on TQ-T3D $\rightarrow$ X task.  On the TQ-DuraGel $\rightarrow$ X task, OmniVaT outperforms SPG \cite{bai2024soft} which relies on the  generative model to obtain class-invariant features by 4.7\% in accuracy with pre-trained ViT-B/16. Under the ViT-L/14 setting, our OmniVaT achieved the best average accuracy and Macro-F1 across 8 domains, reaching 66.9\% and 65.5\%, respectively. The above experimental results demonstrate that our OmniVaT can effectively mitigate modality and domain gaps, achieving efficient single domain generalization for VTL.

To further evaluate the effectiveness of different methods  on the TAG $\rightarrow$ X task, we randomly sample 657,000, 657,000, 308,025, 499,500, and 499,500 intra-class/inter-class pairs from the unseen OF Real, OF 2.0 A, TQ-T3D, [OF Real; TQ-T3D], and [OF 2.0 A; TQ-T3D] domains, respectively, and report the cosine similarity margin \cite{chen2024frequency}, as shown in \cref{T-SM}. Here, [; ] denotes the concatenation of two unseen domains.   Our OmniVaT achieves the largest cosine-similarity margins, reaching 0.191 and 0.143  on the unseen [OF Real; TQ-T3D] and [OF 2.0 A; TQ-T3D] domains, respectively.  These results demonstrate that our  OmniVaT effectively reduces modality and domain gaps across unseen domains, thereby enhancing cross-domain generalization.

 \textit{\textbf{2) Computational Cost:}}
\cref{T-CC} summarizes the trainable parameters (M), training time (s/epoch) on the TAG dataset, GPU memory (MB), inference speed (measured in frames per second with a batch size of 16, FPS),   accuracy (ACC, \%), and  Macro-F1 (F1*, \%) on the single domain generalization TAG $\rightarrow$ X task  for each approach.  As shown in \cref{T-CC}, OmniVaT reduces the training time by 23,038.3 seconds compared with SPG \cite{bai2024soft}, while improving the accuracy from 49.4\% to 56.2\%. Notably, the TAC sensor usually captures tactile images at 30 FPS and OmniVaT reaches an inference speed of 75.2 FPS, demonstrating its suitability for real-time deployment. To assess OmniVaT’s performance in resource-limited conditions, we reduce the representation dimensionality from 512 to 64. The lightweight OmniVaT$^{\ddag}$ not only reduces the number of trainable parameters by 0.1M, 4.5M, 4.8M, and 3.2M compared to PromptStyler \cite{cho2023promptstyler}, SPG \cite{bai2024soft}, MMRL \cite{guo2025mmrl}, and LDC \cite{li2025logits}, respectively, but also achieves the impressive 54.1\% Macro-F1 on the TAG $\rightarrow$ X task. These experiments demonstrate that our OmniVaT effectively achieves a compelling balance between computational efficiency and performance.

\subsection{Ablation Studies}

      \begin{table}[!t]
                  \caption{The influence of the multimodal VIS-TAC image representation learning on the TAG $\rightarrow$ X task. The metrics are accuracy (ACC, \%) and Macro-F1 (F1*, \%).}
  \begin{center}

    \resizebox{0.46\textwidth}{!}{
    \begin{tabular}{c | c  |c |c   }
    \hline   
    \multirow{3}*{Methods}&\multicolumn{1}{c|}{Training Modalities}&\multicolumn{1}{c|}{Test Modalities}& \multicolumn{1}{c}{TAG $\rightarrow$ X}\\
        \cline{2-4}
   ~ &    \multicolumn{3}{c}{\textit{ViT-B / 16 \cite{dosovitskiy2020image} with pre-trained weights  from CLIP \cite{radford2021learning}}} \\
        \cline{2-4}
    ~  &LANG / VIS / TAC&VIS / TAC&ACC / F1*   \\
        \cline{2-4}
                \hline
          \multirow{4}*{MMRL \cite{guo2025mmrl}} &- / $\checkmark$ / - &$\checkmark$ / -& 14.4 / 5.5 \\
            ~& - / $\checkmark$ / $\checkmark$ &$\checkmark$ / $\checkmark$& 14.3 / 5.5 \\
      ~&$\checkmark$ / $\checkmark$ / - &$\checkmark$ / -& 51.0 / 42.6\\
      ~&$\checkmark$ / $\checkmark$ / $\checkmark$&$\checkmark$ / $\checkmark$&46.0 / 33.5\\
                \hline
      \multirow{4}*{LDC \cite{li2025logits}} &- / $\checkmark$ / - &$\checkmark$ / -& 46.8 / 34.4 \\
        ~& - / $\checkmark$ / $\checkmark$ &$\checkmark$ / $\checkmark$& 42.5 / 26.7 \\
      ~&$\checkmark$ / $\checkmark$ / - &$\checkmark$ / -& 52.9 / 43.5\\
      ~&$\checkmark$ / $\checkmark$ / $\checkmark$ &$\checkmark$ / $\checkmark$&50.1 / 36.3\\
          \hline
      \multirow{4}*{OmniVaT (Ours)} &- / $\checkmark$ / - &$\checkmark$ / -& 53.0 / 48.1 \\
      ~& - / $\checkmark$ / $\checkmark$  &$\checkmark$ / $\checkmark$ & 54.5 / 48.0 \\
      ~&$\checkmark$ / $\checkmark$ / - &$\checkmark$ / -& 54.9 / 51.8 \\
      ~&$\checkmark$ / $\checkmark$ / $\checkmark$ &$\checkmark$ / $\checkmark$& 56.2 / 53.7\\
  \hline 
    \end{tabular}
    }
        \label{T-MM}
    \end{center}
    \end{table}

 \textit{\textbf{1) Effectiveness of Multimodal VTL:}} As shown in \cref{T-MM}, compared to single VIS modality generalization, our OmniVaT effectively reduces the  modality discrepancy and improves generalization performance by 1.9\% Macro-F1 on the generalization tasks of TAG $\rightarrow$ X with training LANG information. These results indicate that the proposed OmniVaT effectively mitigates the discrepancies between different modalities and improves domain generalization. Although previous vision-language methods  \cite{guo2025mmrl, li2025logits} achieve good generalization with a single VIS modality, they struggle to handle the  VIS-TAC modality discrepancy, leading to a decline in generalization performance. 

 \textit{\textbf{2) Effectiveness of Key Components:}} As  shown in  \cref{T-Key}, we evaluate the effectiveness of the key components of  the proposed OmniVaT, which is composed of an MFFA module and a DTG module.   We train the proposed modules on  the TAG $\rightarrow$ X and OF 2.0 A  $\rightarrow$ X tasks, respectively. Method 1 represents the Baseline \cite{cho2023promptstyler} method with pre-trained ViT-B/16  \cite{dosovitskiy2020image} weights from CLIP \cite{radford2021learning}. By incorporating MFFA without the MMA loss (+MFFA w/o $\mathcal{L}_{\rm MMA}$), Method 2 achieves a 10.6\% Macro-F1 improvement over the Baseline on the TAG $\rightarrow$ X task, indicating that MFFA enables effective multimodal embedding interactions in a unified embedding-frequency space through FrATT. When the MMA loss is introduced, Method 3 achieves  56.2\% Macro-F1  on the  OF 2.0 A $\rightarrow$ X task, showing that $\mathcal{L}_{\rm  MMA}$ further enhances VIS-TAC alignment and mitigates modality discrepancy. Similarly, Compared with Method 1, adding DTG without the NOD loss (+DTG w/o $\mathcal{L}_{\rm  NOD}$) brings a 1.5\% accuracy increase on the TAG $\rightarrow$X. Method 5 (+DTG) further improves Macro-F1 by 1.8\% over Method 4 on the TAG $\rightarrow$X task, demonstrating that DTG effectively diversifies  fractional representations for single domain generalization. Finally, combining MFFA and DTG (Method 6) yields  +1.7\% accuracy and +2.0\% Macro-F1 over Method 3 on OF 2.0 A $\rightarrow$ X task, confirming that OmniVaT effectively reduces modality discrepancy and enhances domain generalization.

      \begin{table}[!t]
                  \caption{ The influence of key components of the proposed OmniVaT on the TAG $\rightarrow$ X and OF 2.0 A  $\rightarrow$ X tasks, respectively. The metrics are accuracy (ACC, \%) and Macro-F1 (F1*, \%).}
  \begin{center}

    \resizebox{0.46\textwidth}{!}{
    \begin{tabular}{c | l | c | c   }
    \hline   
    \multirow{2}*{\#}&\multirow{2}*{Methods}& \multicolumn{1}{c|}{TAG $\rightarrow$ X}& \multicolumn{1}{c}{OF 2.0 A $\rightarrow$ X}\\
    \cline{3-4}
    ~& ~ &   ACC / {F1*} &ACC / {F1*} \\
    \hline  
      1 & Baseline + $\mathcal{L}_{\rm CE}$& 51.7 / 40.6 &48.9 / 41.2\\
      \hline
      2 & +MFFA (w/o  $\mathcal{L}_{\rm MMA}$) &54.5 / 51.2 &51.4 / 54.6\\
      3 & +MFFA &54.8 / 52.5 &56.9 / 56.2\\

      4 & +DTG (w/o $\mathcal{L}_{\rm NOD}$)&53.2 / 48.0 &53.8 / 52.1\\
      5 & +DTG &53.6 / 49.8 &54.9 / 54.6\\
      6 & +MFFA+DTG &56.2 / 53.7 &58.6 / 58.2\\

    \hline 
    \end{tabular}
    }

        \label{T-Key}
    \end{center}
    \end{table}

       \begin{figure}[t]
    \centering
    \includegraphics[width=0.91\linewidth]{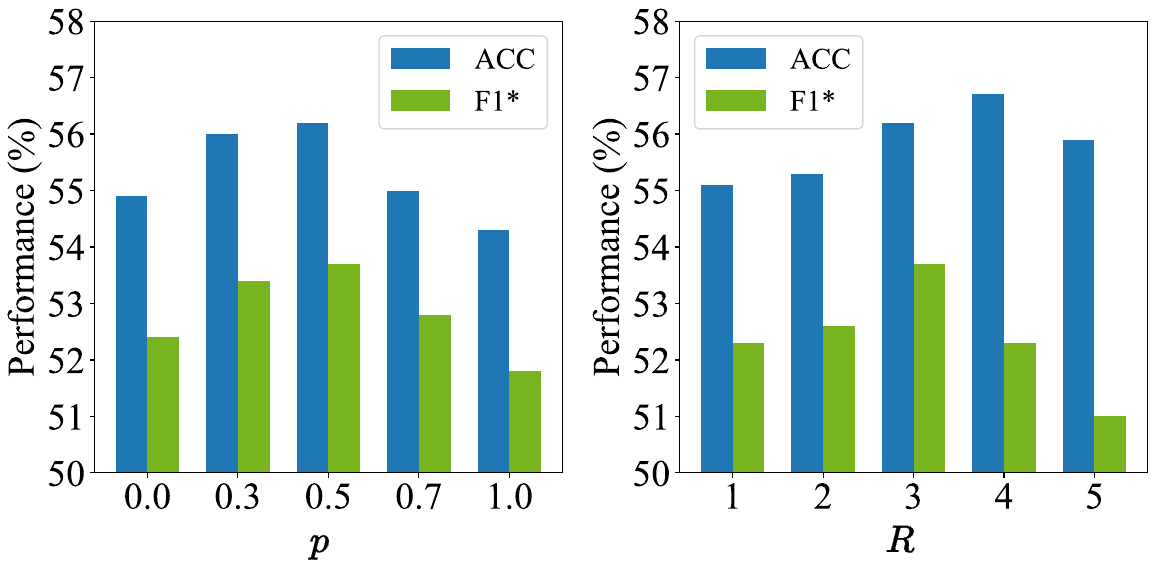}
    \caption{The influence of the fractional order $p$ and the depth $R$ of  tree on the TAG $\rightarrow$ X task. When $p$ = 0.0  or 1.0, it is fixed; otherwise, it is learnable.
}
    \label{FIG_Para}
    \end{figure}
    
\textit{\textbf{3) Effectiveness of the Tree Generation:}}  To evaluate the influence of our discrete tree generation (DTG),  we replace the DTG module in the OmniVaT with different feature generation methods to generate seven feature nodes, as shown in \cref{T-generation}.  Compared with the linear interpolation \cite{ko2020embedding},  series generation, and parallel generation \cite{zhang2023diverse}, our OmniVaT improves at least 1.4\% Macro-F1 on the TAG $\rightarrow$ X task. These results indicate that the discrete tree generation can effectively obtain diverse and reliable multimodal fractional representations  to reduce the domain gaps.

       \begin{table}[!t]
                  \caption{ The influence of the different feature generation methods on the source TAG and OF 2.0 A domains, respectively. The metrics are accuracy (ACC, \%) and Macro-F1 (F1*, \%).}
  \begin{center}

    \resizebox{0.4\textwidth}{!}{
    \begin{tabular}{ l | c | c   }
    \hline   
\multirow{2}*{Methods}& \multicolumn{1}{c|}{TAG $\rightarrow$ X}& \multicolumn{1}{c}{OF 2.0 A $\rightarrow$ X}\\
    \cline{2-3}
     ~ &   ACC / {F1*} &ACC / {F1*} \\
    \hline  
      Baseline & 51.7 / 40.6 &48.9 / 41.2\\
      \hline
     OmniVaT (linear interpolation) &54.6 / 51.5&56.5 / 56.2 \\
     OmniVaT (series generation)& 55.3 / 52.3 &56.9 / 56.6\\
     OmniVaT (parallel generation)& 55.0 / 51.8&57.3 / 57.4\\
     OmniVaT (DTG)&56.2 / 53.7 & 58.6 / 58.2\\
    \hline 
    \end{tabular}
    }
        \label{T-generation}
    \end{center}
    \end{table}

     \textit{\textbf{4) Hyperparameters:}}  We evaluate the influence of  the fractional order $p$ and  depth $R$ of  tree, as shown in \cref{FIG_Para}.  We can observe that compared with the original embedding space ($p=0.0$) and the frequency space ($p=1.0$), the unified fractional domain space ($p=0.5$) achieves more effective single domain generalization for multimodal VTL. In addition, when the depth of  tree $R$ is set to 3, OmniVaT achieves the best overall performance.

\section{Conclusion}
In this work, we observe that VTL inherently suffers from both modality and domain gaps, and we propose a new SDG-VTL task towards real-world applications.  To address the SDG-VTL task, we propose the first OmniVaT framework for the single VIS-TAC domain generalization. OmniVaT integrates an MFFA module  to align VIS-TAC representations within a unified embedding-frequency space, and a DTG module to construct reliable hierarchical fractional features, thus effectively mitigating modality and domain gaps.   Experimental results show OmniVaT outperforms SOTA methods in various single domain generalization for   multimodal VTL scenarios, demonstrating its effectiveness and robustness.  OmniVaT  provides a generic blueprint for extending  single domain generalization beyond the visual modality to broader multimodal learning. Future work will extend OmniVaT to handle vibrotactile data to overcome the common shortcoming of the existing image-based VTL methods for more comprehensive object understanding.

{\small
\bibliographystyle{ieee_fullname}
\bibliography{main}

@article{chen2024frequency,
  title={Frequency-aware feature fusion for dense image prediction},
  author={Chen, Linwei and Fu, Ying and Gu, Lin and Yan, Chenggang and Harada, Tatsuya and Huang, Gao},
  journal={IEEE Transactions on Pattern Analysis and Machine Intelligence},
  year={2024},
}

@inproceedings{loshchilov2017sgdr,
  title={Sgdr: Stochastic gradient descent with warm restarts},
  author={Loshchilov, Ilya and Hutter, Frank},
booktitle={Proceedings of the International Conference on Learning Representations},
  year={2017}
}

@article{zhao2022cross,
  title={Cross-domain classification of multisource remote sensing data using fractional fusion and spatial-spectral domain adaptation},
  author={Zhao, Xudong and Zhang, Mengmeng and Tao, Ran and Li, Wei and Liao, Wenzhi and Philips, Wilfried},
  journal={IEEE Journal of Selected Topics in Applied Earth Observations and Remote Sensing},
  volume={15},
  pages={5721--5733},
  year={2022},
}

@article{candan2000discrete,
  title={The discrete fractional Fourier transform},
  author={Candan, Cagatay and Kutay, M Alper and Ozaktas, Haldun M},
  journal={IEEE Transactions on Signal Processing},
  volume={48},
  number={5},
  pages={1329--1337},
  year={2000},
}

@inproceedings{calli2015ycb,
  title={The ycb object and model set: Towards common benchmarks for manipulation research},
  author={Calli, Berk and Singh, Arjun and Walsman, Aaron and Srinivasa, Siddhartha and Abbeel, Pieter and Dollar, Aaron M},
  booktitle={Proceedings of the International Conference on Advanced Robotics},
  pages={510--517},
  year={2015},
}

@article{lambeta2020digit,
  title={Digit: A novel design for a low-cost compact high-resolution tactile sensor with application to in-hand manipulation},
  author={Lambeta, Mike and Chou, Po-Wei and Tian, Stephen and Yang, Brian and Maloon, Benjamin and Most, Victoria Rose and Stroud, Dave and Santos, Raymond and Byagowi, Ahmad and Kammerer, Gregg and others},
  journal={IEEE Robotics and Automation Letters},
  volume={5},
  number={3},
  pages={3838--3845},
  year={2020},
}

@article{yuan2017gelsight,
  title={Gelsight: High-resolution robot tactile sensors for estimating geometry and force},
  author={Yuan, Wenzhen and Dong, Siyuan and Adelson, Edward H},
  journal={Sensors},
  volume={17},
  number={12},
  pages={2762},
  year={2017},
}

@misc{
  gelsight,
author="GelSight. Inc.",
title="{GelSight Mini}",
url="https://www.gelsight.com/gelsightmini/",
year="2022"
}

@inproceedings{donlon2018gelslim,
  title={Gelslim: A high-resolution, compact, robust, and calibrated tactile-sensing finger},
  author={Donlon, Elliott and Dong, Siyuan and Liu, Melody and Li, Jianhua and Adelson, Edward and Rodriguez, Alberto},
  booktitle={Proceedings of  the IEEE/RSJ International Conference on Intelligent Robots and Systems},
  pages={1927--1934},
  year={2018},
  organization={IEEE}
}

@article{zhang2022tac3d,
  title={Tac3D: A novel vision-based tactile sensor for measuring forces distribution and estimating friction coefficient distribution},
  author={Zhang, Lunwei and Wang, Yue and Jiang, Yao},
  journal={arXiv preprint arXiv:2202.06211},
  year={2022}
}

@article{zhang2024compact,
  title={A compact visuo-tactile robotic skin for micron-level tactile perception},
  author={Zhang, Shixin and Yang, Yiyong and Sun, Fuchun and Bao, Lei and Shan, Jianhua and Gao, Yuan and Fang, Bin},
  journal={IEEE Sensors Journal},
  volume={24},
  number={9},
  pages={15273--15282},
  year={2024},
  publisher={IEEE}
}

@inproceedings{yang2025discrete,
title={Discrete Distribution Networks},
author={Lei Yang},
booktitle={Proceedings of the International Conference on Learning Representations},
year={2025},
}

@article{yao2023tree,
  title={Tree of thoughts: Deliberate problem solving with large language models},
  author={Yao, Shunyu and Yu, Dian and Zhao, Jeffrey and Shafran, Izhak and Griffiths, Tom and Cao, Yuan and Narasimhan, Karthik},
  journal={Proceedings of the Advances in Neural Information Processing Systems},
  volume={36},
  pages={11809--11822},
  year={2023}
}

@inproceedings{miao2024specinfer,
  title={Specinfer: Accelerating large language model serving with tree-based speculative inference and verification},
  author={Miao, Xupeng and Oliaro, Gabriele and Zhang, Zhihao and Cheng, Xinhao and Wang, Zeyu and Zhang, Zhengxin and Wong, Rae Ying Yee and Zhu, Alan and Yang, Lijie and Shi, Xiaoxiang and others},
  booktitle={Proceedings of the ACM International Conference on Architectural Support for Programming Languages and Operating Systems},
  pages={932--949},
  year={2024}
}

@inproceedings{li2025logits,
  title={Logits DeConfusion with CLIP for Few-Shot Learning},
  author={Li, Shuo and Liu, Fang and Hao, Zehua and Wang, Xinyi and Li, Lingling and Liu, Xu and Chen, Puhua and Ma, Wenping},
  booktitle={Proceedings of the IEEE Conference on Computer Vision and Pattern Recognition Conference},
  pages={25411--25421},
  year={2025}
}

@inproceedings{guo2025mmrl,
      title={Mmrl: Multi-modal representation learning for vision-language models},
      author={Guo, Yuncheng and Gu, Xiaodong},
      booktitle={Proceedings of the IEEE Conference on Computer Vision and Pattern Recognition Conference},
      pages={25015--25025},
      year={2025}
}

@inproceedings{RobustVisH2025,
  title={RobustVisH: Robust Visual-Haptic Cross-Modal Recognition Under Transmission Interference},
  author={Zhang, Rouqi and Lu, Chengdi and Lu, Hancheng and Cao, Yang and Zhao, Tiesong},
  booktitle={Proceedings of the  ACM International Conference on Multimedia},
  year={2025},

}

@article{zhou2025slam,
  author={Zhou, Youjie and Mei, Guofeng and Wang, Yiming and Wan, Yi and Poiesi, Fabio},
  journal={IEEE Robotics and Automation Letters}, 
  title={Multimodal Fusion SLAM With Fourier Attention}, 
  year={2025},
  volume={10},
  number={2},
  pages={1050-1057},
}

@article{chen2025bilinear,
  title={Bilinear Parallel Fourier Transformer for Multimodal Remote Sensing Classification},
  author={Chen, Yaxiong and Wang, Qicong and Zhao, Yichen and Xiong, Shengwu and Lu, Xiaoqiang},
  journal={IEEE Transactions on Geoscience and Remote Sensing},
  year={2025},
  publisher={IEEE}
}

@article{oquab2023dinov2,
  title={Dinov2: Learning robust visual features without supervision},
  author={Oquab, Maxime and Darcet, Timoth{\'e}e and Moutakanni, Th{\'e}o and Vo, Huy and Szafraniec, Marc and Khalidov, Vasil and Fernandez, Pierre and Haziza, Daniel and Massa, Francisco and El-Nouby, Alaaeldin and others},
  journal={arXiv preprint arXiv:2304.07193},
  year={2023}
}

@article{zhang2025vtla,
  title={Vtla: Vision-tactile-language-action model with preference learning for insertion manipulation},
  author={Zhang, Chaofan and Hao, Peng and Cao, Xiaoge and Hao, Xiaoshuai and Cui, Shaowei and Wang, Shuo},
  journal={arXiv preprint arXiv:2505.09577},
  year={2025}
}

@inproceedings{bai2024soft,
  title={Soft prompt generation for domain generalization},
  author={Bai, Shuanghao and Zhang, Yuedi and Zhou, Wanqi and Luan, Zhirong and Chen, Badong},
  booktitle={Proceedings of the  European Conference on Computer Vision},
  pages={434--450},
  year={2024},
}

@inproceedings{he2016deep,
  title={Deep residual learning for image recognition},
  author={He, Kaiming and Zhang, Xiangyu and Ren, Shaoqing and Sun, Jian},
  booktitle={Proceedings of the IEEE Conference on Computer Vision and Pattern Recognition},
  pages={770--778},
  year={2016}
}

@ARTICLE{zhang2025frequency,
  author={Zhang, Yukang and Wang, Hanzi and Lu, Yang and Yan, Yan and Li, Xuelong},
  journal={IEEE Transactions on Information Forensics and Security}, 
  title={Frequency Domain Nuances Mining for Visible-Infrared Person Re-Identification}, 
  year={2025},
  volume={20},
  pages={5411-5424},
}

@inproceedings{ko2020embedding,
  title={Embedding expansion: Augmentation in embedding space for deep metric learning},
  author={Ko, Byungsoo and Gu, Geonmo},
  booktitle={Proceedings of the IEEE Conference on Computer Vision and Pattern Recognition},
  pages={7255--7264},
  year={2020}
}

@inproceedings{zhang2023diverse,
  title={Diverse embedding expansion network and low-light cross-modality benchmark for visible-infrared person re-identification},
  author={Zhang, Yukang and Wang, Hanzi},
  booktitle={Proceedings of the IEEE Conference on Computer Vision and Pattern Recognition},
  pages={2153--2162},
  year={2023}
}

@article{yu2023deep,
  title={Deep fractional Fourier transform},
  author={Yu, Hu and Huang, Jie and Li, Lingzhi and Zhao, Feng and others},
  journal={Proceedings of the  Advances in Neural Information Processing Systems},
  volume={36},
  pages={72761--72773},
  year={2023}
}

@article{chen2024fractional,
  title={Fractional Fourier Based Frequency-Spatial-Spectral Prototype Network for Agricultural Hyperspectral Image Open-Set Classification},
  author={Chen, Maoyang and Feng, Shou and Zhao, Chunhui and Qu, Bo and Su, Nan and Li, Wei and Tao, Ran},
  journal={IEEE Transactions on Geoscience and Remote Sensing},
  year={2024},
  publisher={IEEE}
}

@article{kocc2024trainable,
  title={Trainable fractional Fourier transform},
  author={Ko{\c{c}}, Emirhan and Alika{\c{s}}ifo{\u{g}}lu, Tuna and Aras, Arda Can and Ko{\c{c}}, Aykut},
  journal={IEEE Signal Processing Letters},
  volume={31},
  pages={751--755},
  year={2024},
  publisher={IEEE}
}

@article{zhao2022fractional,
  title={Fractional Fourier image transformer for multimodal remote sensing data classification},
  author={Zhao, Xudong and Zhang, Mengmeng and Tao, Ran and Li, Wei and Liao, Wenzhi and Tian, Lianfang and Philips, Wilfried},
  journal={IEEE Transactions on Neural Networks and Learning Systems},
  volume={35},
  number={2},
  pages={2314--2326},
  year={2022},
  publisher={IEEE}
}

@article{ji2025frequency,
  title={Frequency-Spatial Complementation: Unified Channel-Specific Style Attack for Cross-Domain Few-Shot Learning},
  author={Ji, Zhong and Wang, Zhilong and Liu, Xiyao and Yu, Yunlong and Pang, Yanwei and Han, Jungong},
  journal={IEEE Transactions on Image Processing},
  year={2025},
  publisher={IEEE}
}

@inproceedings{gunduboina2025frogdognet,
  title={FrogDogNet: Fourier frequency Retained visual prompt Output Guidance for Domain Generalization of CLIP in Remote Sensing},
  author={Gunduboina, Hariseetharam and Khan, Muhammad Haris and Banerjee, Biplab},
  booktitle={Proceedings of the IEEE Conference on Computer Vision and Pattern Recognition Conference},
  pages={2359--2372},
  year={2025}
}

@article{lu2022domain,
  title={Domain-invariant feature exploration for domain generalization},
  author={Lu, Wang and Wang, Jindong and Li, Haoliang and Chen, Yiqiang and Xie, Xing},
  journal={Transactions on Machine Learning Research},
  year={2022}
}

@inproceedings{xu2021fourier,
  title={A fourier-based framework for domain generalization},
  author={Xu, Qinwei and Zhang, Ruipeng and Zhang, Ya and Wang, Yanfeng and Tian, Qi},
  booktitle={Proceedings of the IEEE Conference on Computer Vision and Pattern Recognition},
  pages={14383--14392},
  year={2021}
}

@ARTICLE{si2022taxim,
  author={Si, Zilin and Yuan, Wenzhen},
  journal={IEEE Robotics and Automation Letters}, 
  title={Taxim: An Example-Based Simulation Model for GelSight Tactile Sensors}, 
  year={2022},
  volume={7},
  number={2},
  pages={2361-2368},
}

@inproceedings{gao2022objectfolder,
  title={Objectfolder 2.0: A multisensory object dataset for sim2real transfer},
  author={Gao, Ruohan and Si, Zilin and Chang, Yen-Yu and Clarke, Samuel and Bohg, Jeannette and Fei-Fei, Li and Yuan, Wenzhen and Wu, Jiajun},
  booktitle={Proceedings of the IEEE Conference on Computer Vision and Pattern Recognition},
  pages={10598--10608},
  year={2022}
}

@inproceedings{yang2022touch,
  title={Touch and go: Learning from human-collected vision and touch},
  author={Yang, Fengyu and Ma, Chenyang and Zhang, Jiacheng and Zhu, Jing and Yuan, Wenzhen and Owens, Andrew},
  booktitle={Proceedings of the Advances in Neural Information Processing Systems},
  year={2022}
}

@ inproceedings{dave2024multimodal,
  author={Dave, Vedant and Lygerakis, Fotios and Rueckert, Elmar},
  booktitle={IEEE International Conference on Robotics and Automation}, 
  title={Multimodal Visual-Tactile Representation Learning through Self-Supervised Contrastive Pre-Training}, 
  year={2024},
  volume={},
  number={},
  pages={8013-8020},
}

@inproceedings{yang2024binding,
  title={Binding touch to everything: Learning unified multimodal tactile representations},
  author={Yang, Fengyu and Feng, Chao and Chen, Ziyang and Park, Hyoungseob and Wang, Daniel and Dou, Yiming and Zeng, Ziyao and Chen, Xien and Gangopadhyay, Rit and Owens, Andrew and others},
  booktitle={Proceedings of the IEEE Conference on Computer Vision and Pattern Recognition},
  pages={26340--26353},
  year={2024}
}

@article{li2023vito,
  title={VITO-transformer: a visual-tactile fusion network for object recognition},
  author={Li, Baojiang and Bai, Jibo and Qiu, Shengjie and Wang, Haiyan and Guo, Yuting},
  journal={IEEE Transactions on Instrumentation and Measurement},
  year={2023},
}

@inproceedings{wei2022alignment,
  title={Alignment and Multi-Scale Fusion for Visual-Tactile Object Recognition},
  author={Wei, Fuyang and Zhao, Jianhui and Shan, Chudong and Yuan, Zhiyong},
  booktitle={Proceedings of the IEEE International Joint Conference on Neural Networks},
  pages={1--8},
  year={2022},
}

@inproceedings{gao2023objectfolder,
  title={The objectfolder benchmark: Multisensory learning with neural and real objects},
  author={Gao, Ruohan and Dou, Yiming and Li, Hao and Agarwal, Tanmay and Bohg, Jeannette and Li, Yunzhu and Fei-Fei, Li and Wu, Jiajun},
  booktitle={Proceedings of the IEEE Conference on Computer Vision and Pattern Recognition},
  pages={17276--17286},
  year={2023}
}

@inproceedings{feng2025anytouch,
  title={Anytouch: Learning unified static-dynamic representation across multiple visuo-tactile sensors},
  author={Feng, Ruoxuan and Hu, Jiangyu and Xia, Wenke and Gao, Tianci and Shen, Ao and Sun, Yuhao and Fang, Bin and Hu, Di},
  booktitle ={Proceedings of the International Conference on Learning Representations},
  year={2025}
}

@inproceedings{cho2023promptstyler,
  title={Promptstyler: Prompt-driven style generation for source-free domain generalization},
  author={Cho, Junhyeong and Nam, Gilhyun and Kim, Sungyeon and Yang, Hunmin and Kwak, Suha},
  booktitle={Proceedings of the IEEE International Conference on Computer Vision},
  pages={15702--15712},
  year={2023}
}

@article{xu2025batstyler,
  title={BatStyler: Advancing Multi-category Style Generation for Source-free Domain Generalization},
  author={Xu, Xiusheng and Qi, Lei and Zhou, Jingyang and Geng, Xin},
  journal={IEEE Transactions on Circuits and Systems for Video Technology},
  year={2025},
}

@inproceedings{zhang2025promptta,
  title={Promptta: Prompt-driven text adapter for source-free domain generalization},
  author={Zhang, Haoran and Bai, Shuanghao and Zhou, Wanqi and Fu, Jingwen and Chen, Badong},
  booktitle={Proceedings of the IEEE International Conference on Acoustics, Speech and Signal Processing},
  pages={1--5},
  year={2025},
}

@article{liu2024stydesty,
  title={StyDeSty: Min-max stylization and destylization for single domain generalization},
  author={Liu, Songhua and Jin, Xin and Yang, Xingyi and Ye, Jingwen and Wang, Xinchao},
  journal={Proceedings of the International Conference on Machine Learning},
  year={2024}
}

@inproceedings{chen2023meta,
  title={Meta-causal learning for single domain generalization},
  author={Chen, Jin and Gao, Zhi and Wu, Xinxiao and Luo, Jiebo},
  booktitle={Proceedings of the IEEE Conference on Computer Vision and Pattern Recognition},
  pages={7683--7692},
  year={2023}
}

@inproceedings{dosovitskiy2020image,
  title={An image is worth 16x16 words: Transformers for image recognition at scale},
  author={Dosovitskiy, Alexey and Beyer, Lucas and Kolesnikov, Alexander and Weissenborn, Dirk and Zhai, Xiaohua and Unterthiner, Thomas and Dehghani, Mostafa and Minderer, Matthias and Heigold, Georg and Gelly, Sylvain and others},
  booktitle={Proceedings of the International Conference on Learning Representations},
  year={2020}
}

@inproceedings{zheng2024advst,
  title={AdvST: Revisiting data augmentations for single domain generalization},
  author={Zheng, Guangtao and Huai, Mengdi and Zhang, Aidong},
  booktitle={Proceedings of the AAAI conference on artificial intelligence},
  pages={21832--21840},
  year={2024}
}

@inproceedings{xu2025adv,
    title     = {Adversarial Domain Prompt Tuning and Generation for Single Domain Generalization},
    author    = {Xu, Zhipeng and Cheng, De and Jiang, Xinyang and Wang, Nannan and Li, Dongsheng and Gao, Xinbo},
    booktitle = {Proceedings of the IEEE Conference on Computer Vision and Pattern Recognition Conference},
    year      = {2025},
    pages     = {18584-18595}
}

@inproceedings{gupta2025sensorinvariant,
  title={Sensor-Invariant Tactile Representation},
  author={Harsh Gupta and Yuchen Mo and Shengmiao Jin and Wenzhen Yuan},
  booktitle={Proceedings of the International Conference on Learning Representations},
  year={2025}
}

@inproceedings{radford2021learning,
  title={Learning transferable visual models from natural language supervision},
  author={Radford, Alec and Kim, Jong Wook and Hallacy, Chris and Ramesh, Aditya and Goh, Gabriel and Agarwal, Sandhini and Sastry, Girish and Askell, Amanda and Mishkin, Pamela and Clark, Jack and others},
  booktitle={Proceedings of the  International Conference on Machine Learning},
  pages={8748--8763},
  year={2021},
}
}

\newpage
\appendix

\section{More Fractional Fourier Transform Details}

The $p$-th order Fractional Fourier Transform (FrFT) of the  embedding $\mathbf{E} \in \mathbb{R}^{1 \times D}$, is defined as follows:
\begin{equation}
{\rm FrFT}_{p}(\mathbf{E}(u_0)) = \int_{-\infty}^{\infty} K_p(u_0, u_p) \mathbf{E}(u_0) \, du_0,
\label{Eq-FrFT}
\end{equation}
where $K_p(u_0, u_p)$ is the kernel function of the fractional Fourier transform, which can be seen in \cref{{T-K}}, $u_0$ and $u_p$  represent the original embedding space and  fractional space, respectively.   Although \cref{Eq-FrFT} defines the continuous FrFT, practical embeddings are discrete signals. Therefore, we employ the discrete FrFT (DFrFT) \cite{candan2000discrete}, that is
\begin{equation}
\mathbf{E}(u_p)[m] = \sum\nolimits_{n=1}^{D} \mathbf{F}_{p}[m,n]\mathbf{E}(u_0)[n], m={1,\cdots, D}
\end{equation}
where $\mathbf{F}_{p}[m,n]= \sum_{k=1}^{D} v_k(m)\, (\lambda_k)^{p}\, v_k(n)$ is the discrete FrFT matrix \cite{candan2000discrete},  $v_k$ is the $k$-th discrete Hermite-Gaussians vector, and $\lambda_k$ represents the corresponding eigenvalues. The final $p$-order discrete FrFT of embedding $\mathbf{E}$ can be achieved as  $\mathbf{E}(u_p)=\mathbf{F}_{p}\mathbf{E}$.

\begin{table}[h]
\centering
\caption{Kernel function $K_p(u_0, u_p)$ of the Fractional Fourier Transform (FrFT) under different $\alpha$ values ($\alpha= p\pi/2$). $A_{\alpha}=\sqrt{(1 - j\cot\alpha)/2\pi}$ and $\delta(\cdot)$ is the Dirac delta function.}
\resizebox{0.5\textwidth}{!}{\begin{tabular}{c|c}
\hline
 {Kernel Expression} $K_p(u_0, u_p)$& {Condition} \\
 \hline
 $A_{\alpha}\exp{j\left(u_0^2 \cot \alpha /2 - u_p u_0 \csc \alpha + u_p^2 \cot \alpha /2\right)}$ & $\alpha \neq n\pi$ \\

$\delta(u_0 - u_p)$ &$ \alpha = 2n\pi$ \\

$\delta(u_0 + u_p)$ &$\alpha= (2n + 1)\pi$ \\
 \hline
\end{tabular}}
\label{T-K}
\end{table}

\section{More OmniVaT Details}

The  training and  inference processes of the proposed OmniVaT are summarized in \cref{alg-train,alg-test}, respectively.

\begin{algorithm}[t]
\caption{Training process of OmniVaT.}
\label{alg-train}
\begin{algorithmic}[1]
\STATE \textbf{Input:}  Single source VIS-TAC domain ${\mathcal{S}}=\{x^{v}, x^{t},  y\}$ with  auxiliary LANG  information ${x}^{l}$; initial parameters $\theta$ of OmniVaT; epochs $H$; mini-batch size $B$.
\STATE \textbf{Output:} The final  parameters $\theta^{\prime}$ of OmniVaT.
\FOR{$h = 1$ to $H$}
    \FOR{each mini-batch $\{x^{v}, x^{t}, x^{l}\}_b^B$ in $\mathcal{S}$}
        \STATE Obtain the VIS, TAC, and LANG  embeddings by the frozen multimodal encoders.
	 \STATE Apply the  MFFA module to map multimodal VIS, TAC, and LANG  embeddings  into the unified embedding-frequency space to mitigate the modality gap via Eqs. (3-6).

	\STATE Apply the  DTG module to obtain the diverse and reliable  features $\mathbf{T}^{tree}$  for improving the  domain generalization ability of the model.
     
\STATE Compute the joint loss via Eq. (12).

         \STATE Update $\theta$ by stochastic gradient descent.
    \ENDFOR
\ENDFOR
\end{algorithmic}
\end{algorithm}

\begin{algorithm}[t]
\caption{Inference process of OmniVaT.}
\label{alg-test}
\begin{algorithmic}[1]
\STATE \textbf{Input:}  Unseen target VIS-TAC domains ${\mathcal{T}}=\{x^{v}, x^{t}\}$; the final parameters $\theta^{\prime}$ of OmniVaT.\\
\STATE \textbf{Output:} The predicted class of the unseen VIS–TAC sample.\\
    \FOR{each  $\{x^{v}, x^{t}\}$ in $\mathcal{T}$}
        \STATE Obtain the VIS  and TAC  embeddings by the frozen multimodal encoders.
	 \STATE Apply the  MFFA module to map multimodal VIS and TAC embeddings into the unified embedding-frequency space and improve cross-modal semantic consistency.

\STATE  Compute the class probability from the multimodal fractional features.

    \ENDFOR
\end{algorithmic}
\end{algorithm}

\section{More Dataset and Task Details}

On the Touch-and-Go (TAG) dataset \cite{yang2022touch}, TAC data are captured by the GelSight sensor \cite{yuan2017gelsight}. The OF Real dataset \cite{gao2023objectfolder} uses the GelSlim sensor \cite{donlon2018gelslim}, while OF 2.0 \cite{gao2022objectfolder} provides simulated tactile signals generated by Taxim \cite{si2022taxim}. We denote the first 100 objects in the OF 2.0 dataset as OF 2.0 A, which are sourced from 3D Model Haven, YCB \cite{calli2015ycb}, and Google Scanned Objects, while the remaining objects are referred to as OF 2.0 B. The TacQuad (TQ) dataset \cite{feng2025anytouch} includes four tactile sensors:  DIGIT \cite{lambeta2020digit}, GelSight Mini \cite{gelsight}, DuraGel \cite{zhang2024compact}, and Tac3D \cite{zhang2022tac3d}, yielding the TQ-DIGIT, TQ-GelMini, TQ-DuraGel, and TQ-T3D domains. Dataset statistics are summarized in \cref{T-datasets}. Since OF 2.0 A and OF 2.0 B share the same Taxim-based TAC modality, we do not evaluate across them to avoid domain leakage.  Because TQ domains do not include ceramic objects, we exclude ceramic samples when constructing unseen domains. Furthermore,  when training on TQ domains, we partition it into indoor and outdoor series so that the model is trained and evaluated on non-overlapping data.  The SDG-VTL settings are detailed in \cref{T-Setting}.

\begin{table}[t]
\centering
\caption{Summarization of the datasets (domains). Each dataset (domain) provides VIS-TAC samples.}
\label{T-datasets}
    \resizebox{0.42\textwidth}{!}
{%
\begin{tabular}{rccc}
\hline
 {Domains}&{TAC Sensors}&{Training Set} &{Test Set}\\
\hline
TAG \cite{yang2022touch}     & GelSight & 5,520 & 1,770\\
OF Real \cite{gao2023objectfolder}   & GelSlim & 3,401 & 1,355\\
OF 2.0 A  \cite{gao2022objectfolder}    & Taxim&  1,000& 2,200 \\
OF 2.0 B \cite{gao2022objectfolder}     & Taxim & 9,500& 2,500\\
\hline

\textit{TQ  Indoor-Outdoor  Series}\\
TQ-DIGIT \cite{feng2025anytouch}  & DIGIT & - & 810 \\
TQ-GelMini \cite{feng2025anytouch}  & GelSight Mini & - & 810\\
TQ-DuraGel \cite{feng2025anytouch}  &DuraGel & - & 810\\
\hline
\textit{TQ  Indoor  Series}\\

TQ-DIGIT (Indoor) \cite{feng2025anytouch}  & DIGIT & - & 525 \\
TQ-GelMini  (Indoor) \cite{feng2025anytouch}  & GelSight Mini & - & 525\\
TQ-DuraGel  (Indoor) \cite{feng2025anytouch}  &DuraGel & - & 525\\
TQ-T3D  (Indoor) \cite{feng2025anytouch}  & Tac3D & 1,196 & 682\\
\hline
\textit{TQ  Outdoor  Series}\\
TQ-DIGIT (Outdoor) \cite{feng2025anytouch}  & DIGIT & 779 & 285 \\
TQ-GelMini  (Outdoor) \cite{feng2025anytouch}  & GelSight Mini & 779 & 285\\
TQ-DuraGel  (Outdoor) \cite{feng2025anytouch}  &DuraGel & 779 & 285\\
\hline
\end{tabular}
}
\end{table}

\begin{table}[t]
\centering
\caption{The settings of the single domain generalization for multimodal VIS-TAC learning (SDG-VTL) task.}
\label{T-Setting}
{
    \resizebox{0.42\textwidth}{!}{\begin{tabular}{cc}
\hline

{Seen Domain}& {Unseen Domains}\\
\hline
TAG \cite{yang2022touch}&\makecell{OF Real \cite{gao2023objectfolder}, OF 2.0 A \cite{gao2022objectfolder},  \\ OF 2.0 B \cite{gao2022objectfolder},  TQ-DIGIT\cite{feng2025anytouch},  \\ TQ-GelMini \cite{feng2025anytouch}, TQ-DuraGel \cite{feng2025anytouch}, TQ-T3D \cite{feng2025anytouch}} \\
\hline
OF Real \cite{gao2023objectfolder} &\makecell{TAG \cite{yang2022touch}, OF 2.0 A \cite{gao2022objectfolder}, \\ OF 2.0 B \cite{gao2022objectfolder},  TQ-DIGIT \cite{feng2025anytouch}, \\TQ-GelMini \cite{feng2025anytouch}, TQ-DuraGel \cite{feng2025anytouch}, TQ-T3D \cite{feng2025anytouch}} \\
\hline

OF 2.0 A \cite{gao2022objectfolder}&\makecell{TAG \cite{yang2022touch}, OF Real \cite{gao2023objectfolder},   TQ-DIGIT \cite{feng2025anytouch}, \\ TQ-GelMini \cite{feng2025anytouch},  TQ-DuraGel \cite{feng2025anytouch}, TQ-T3D \cite{feng2025anytouch}} \\
\hline

OF 2.0 B \cite{gao2022objectfolder}&\makecell{TAG \cite{yang2022touch}, OF Real \cite{gao2023objectfolder},  TQ-DIGIT \cite{feng2025anytouch},  \\ TQ-GelMini \cite{feng2025anytouch},  TQ-DuraGel \cite{feng2025anytouch}, TQ-T3D \cite{feng2025anytouch}} \\
\hline

\makecell{TQ-DIGIT \cite{feng2025anytouch} \\  (Outdoor) }&\makecell{TAG\cite{yang2022touch}, OF Real \cite{gao2023objectfolder}, \\ OF 2.0 A \cite{gao2022objectfolder}, OF 2.0 B \cite{gao2022objectfolder}, TQ-GelMini (Indoor)  \cite{feng2025anytouch} , \\ TQ-DuraGel (Indoor)  \cite{feng2025anytouch} , TQ-T3D (Indoor)  \cite{feng2025anytouch} } \\
\hline

\makecell{TQ-GelMini \cite{feng2025anytouch} \\ (Outdoor)}&\makecell{TAG\cite{yang2022touch}, OF Real \cite{gao2023objectfolder}, \\ OF 2.0 A \cite{gao2022objectfolder}, OF 2.0 B \cite{gao2022objectfolder},   TQ-DIGIT (Indoor) \cite{feng2025anytouch}, \\ TQ-DuraGel (Indoor) \cite{feng2025anytouch}, TQ-T3D (Indoor) \cite{feng2025anytouch}} \\
\hline

\makecell{TQ-DuraGel \cite{feng2025anytouch} \\  (Outdoor)}&\makecell{TAG\cite{yang2022touch}, OF Real \cite{gao2023objectfolder}, \\ OF 2.0 A \cite{gao2022objectfolder}, OF 2.0 B \cite{gao2022objectfolder},  TQ-DIGIT (Indoor) \cite{feng2025anytouch}, \\ TQ-GelMini (Indoor) \cite{feng2025anytouch}, TQ-T3D (Indoor) \cite{feng2025anytouch}} \\
\hline

\makecell{TQ-T3D \cite{feng2025anytouch} \\  (Indoor)}&\makecell{TAG\cite{yang2022touch}, OF Real \cite{gao2023objectfolder}, \\ OF 2.0 A \cite{gao2022objectfolder}, OF 2.0 B \cite{gao2022objectfolder},  TQ-DIGIT (Outdoor) \cite{feng2025anytouch}, \\TQ-GelMini (Outdoor) \cite{feng2025anytouch}, TQ-DuraGel (Outdoor) \cite{feng2025anytouch}} \\

\hline

\end{tabular}%
}}
\end{table}

\section{More Ablation Studies}

    \textit{\textbf{1) Effectiveness of the Global Class token:}}  As shown in \cref{T-global-cls-token}, the global class token brings  improvements to OmniVaT. Specifically, OmniVaT with the global class token outperforms its counterpart without it by 0.2\% and 0.5\% in Macro-F1 on the TAG $\rightarrow$ X and OF 2.0 A $\rightarrow$ X tasks, respectively. This shows that the global class token can reduce intra-class variation and improve SDG-VTL performance.

       \begin{table}[!t]
                  \caption{ The influence of the global class token of the proposed OmniVaT on the source TAG and OF 2.0 A domains, respectively. The metrics are accuracy (ACC, \%) and Macro-F1 (F1*, \%).}
  \begin{center}

    \resizebox{0.43\textwidth}{!}{
    \begin{tabular}{ l | c | c   }
    \hline   
\multirow{2}*{Methods}& \multicolumn{1}{c|}{TAG $\rightarrow$ X}& \multicolumn{1}{c}{OF 2.0 A $\rightarrow$ X}\\
    \cline{2-3}
 ~ &   ACC / {F1*} &ACC / {F1*} \\
    \hline  
 Baseline & 51.7 / 40.6 &48.9 / 41.2\\
      \hline
  OmniVaT (w/o global class token) &55.8 / 53.5&58.3 / 57.7\\
   OmniVaT (w global class token) &56.2 / 53.7 & 58.6 / 58.2\\
    \hline 
    \end{tabular}
    }
        \label{T-global-cls-token}
    \end{center}
    \end{table}

\begin{figure}[t]
    \centering
    \includegraphics[width=0.8 \linewidth]{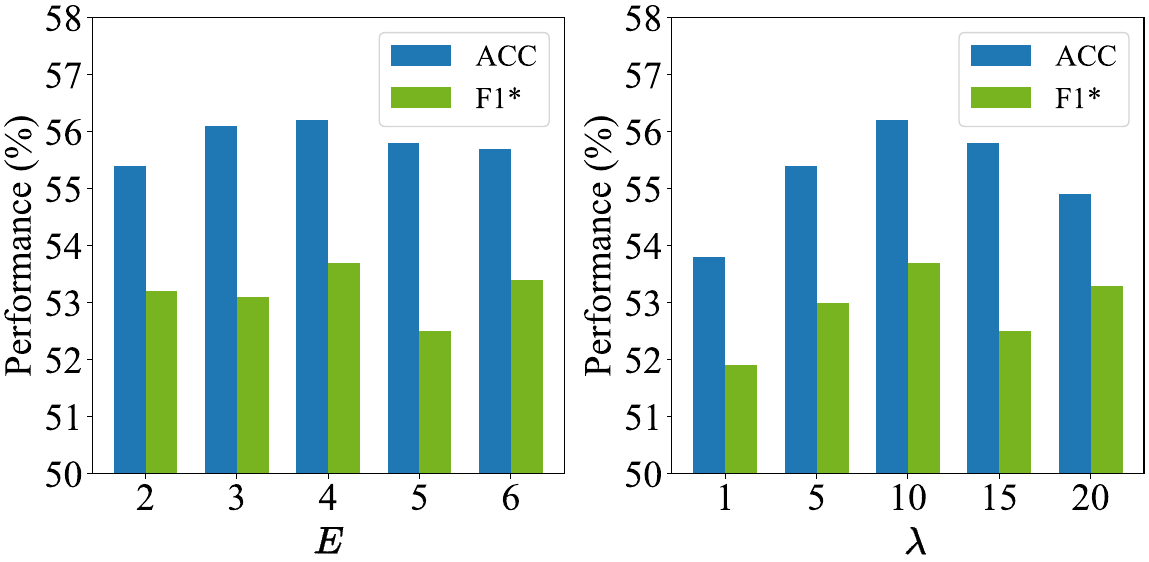}
    \caption{The influence of the extension parameter $E$ and the balance weight $\lambda$  on the TAG $\rightarrow$ X task.}
    \label{FIG_Supp_Para}
    \end{figure}

\textit{\textbf{2) Hyperparameters:}}  \cref{FIG_Supp_Para} illustrates the variations in accuracy and Macro-F1 with respect to the extension parameter $E$ and the balance weight $\lambda$ on the TAG $\rightarrow$ X task. It can be observed that the optimal performance is obtained when $E=4$ and $\lambda=10$.

\section{More Comparison Results}
\cref{T-TAG,T-OFReal,T-OF20A,T-OF20B,T-TQDI,T-TQGEL,T-TQDUR,T-TQ3D} provide detailed comparisons of OmniVaT and prior SOTA methods on 8 VIS-TAC domains. Overall, OmniVaT achieves the highest average performance.  This demonstrates that our OmniVaT is able to achieve more effective single-domain generalization for VIS-TAC learning.    

\begin{table*}[t]
  \centering
  \caption{Comparison with state-of-the-art methods on the TAG $\rightarrow$ X task. “TAG $\rightarrow$ X”  represents the evaluation of the other unseen target domains, excluding the seen source TAG domain. \textcolor{red}{$\Delta$ LDC}  denotes the improvements of the proposed OmniVaT over LDC. The metrics are accuracy (ACC, \%) and Macro-F1 (F1*, \%).  }
   \resizebox{1.0\textwidth}{!}{
  \begin{tabular}{rcccccccc}
   \hline
   
   \hline
  \multirow{2}{*}{Methods  / {\small Venue}} &{\makecell[c]{$\rightarrow$ OF Real}} & {\makecell[c]{$\rightarrow$ OF 2.0 A}} & {\makecell[c]{ $\rightarrow$ OF 2.0 B}}&{\makecell[c]{$\rightarrow$ TQ-DIGIT}}&{\makecell[c]{ $\rightarrow$ TQ-GelMini}} & {\makecell[c]{$\rightarrow$ TQ-DuraGel}}&{\makecell[c]{ $\rightarrow$ TQ-T3D}} &{Average}\\ 
  \cline{2-9}
  ~  & ACC / F1* & ACC / F1* & ACC / F1*& ACC / F1*& ACC / F1*& ACC / F1* & ACC / F1*& ACC / F1* \\ 
        \hline

   ~ &    \multicolumn{8}{c}{ \textit{ViT-B / 16 \cite{dosovitskiy2020image} with pre-trained weights  from CLIP \cite{radford2021learning}}} \\
            \hline

  CLIP  /  {\small ICML-21} \cite{radford2021learning} &49.0 / 48.8 &51.6 / 37.5&48.1 / 46.0 &  37.3 / 32.6 & 47.5 / 31.8&44.4 / 31.8&  56.9 / 51.5&47.8 / 40.0\\
  PromptStyler  /  {\small ICCV-23} \cite{cho2023promptstyler} & 47.5 / 37.4& 46.0 / 24.8& 43.3 / 22.4& 59.3 / 58.9& 53.8 / 46.8& 53.6 / 49.3& 58.4 / 44.4 &51.7 / 40.6\\
  SPG  /  {\small ECCV-24} \cite{bai2024soft} &46.3 / 37.3&39.5 / 19.9&38.4 / 27.2& 52.8 / 52.4&55.9 / 55.5&54.7 / 43.0&58.2 / 57.8&49.4 / 41.9 \\
    MMRL /  {\small CVPR-25} \cite{guo2025mmrl}&37.4 / 24.7&37.5 / 18.0&26.7 / 17.4&53.1 / 46.6&56.0 / 49.1&54.4 / 52.5&57.1 / 26.4 &46.0 / 33.5\\
    
    LDC /  {\small CVPR-25} \cite{li2025logits}&51.2 / 36.0&43.8 / 27.6&41.2 / 34.1&49.4 / 47.6&53.8 / 44.8&54.2 / 33.6&57.1 / 30.1&50.1 / 36.3\\
                 \hline

    OmniVaT /  {\small Ours}&56.7 / 50.8&52.3 / 38.1&48.7 / 46.7&61.3 / 60.0&58.7 / 60.4&56.1 / 59.2&59.3 / 60.5&56.2 / 53.7\\
   
    \textcolor{red}{$\Delta$ LDC}&\textcolor{red}{5.5} / \textcolor{red}{14.8}&\textcolor{red}{8.5} / \textcolor{red}{10.5}&\textcolor{red}{7.5} / \textcolor{red}{12.6}&\textcolor{red}{11.9} / \textcolor{red}{12.4}&\textcolor{red}{4.9} / \textcolor{red}{15.6}&\textcolor{red}{1.9} / \textcolor{red}{25.6}&\textcolor{red}{2.2} / \textcolor{red}{30.4}&\textcolor{red}{6.1} / \textcolor{red}{17.4}\\

      \hline
             
             \hline
  \end{tabular}
  
}
        \label{T-TAG}

\end{table*}

\begin{table*}[t]
  \centering
  \caption{Comparison with state-of-the-art methods on the OF Real $\rightarrow$ X task. “OF Real $\rightarrow$ X”  represents the evaluation of the other unseen target domains, excluding the seen source OF Real domain.  \textcolor{red}{$\Delta$ LDC}  denotes the improvements of the proposed OmniVaT over LDC. The metrics are accuracy (ACC, \%) and Macro-F1 (F1*, \%).  }
   \resizebox{1.0\textwidth}{!}{
  \begin{tabular}{rcccccccc}
   \hline
   
   \hline
  \multirow{2}{*}{Methods  / {\small Venue}} &{\makecell[c]{$\rightarrow$ TAG}} & {\makecell[c]{$\rightarrow$ OF 2.0 A}} & {\makecell[c]{ $\rightarrow$ OF 2.0 B}}&{\makecell[c]{$\rightarrow$ TQ-DIGIT}}&{\makecell[c]{ $\rightarrow$ TQ-GelMini}} & {\makecell[c]{$\rightarrow$ TQ-DuraGel}}&{\makecell[c]{ $\rightarrow$ TQ-T3D}} &{Average}\\ 
  \cline{2-9}
  ~  & ACC / F1* & ACC / F1* & ACC / F1*& ACC / F1*& ACC / F1*& ACC / F1* & ACC / F1*& ACC / F1* \\ 
        \hline

   ~ &    \multicolumn{8}{c}{ \textit{ViT-B / 16 \cite{dosovitskiy2020image} with pre-trained weights  from CLIP \cite{radford2021learning}}} \\
            \hline

  CLIP  /  {\small ICML-21} \cite{radford2021learning} & 51.4 / 49.8&51.6 / 37.5&48.1 / 46.0 &  37.3 / 32.6 & 47.5 / 31.8&44.4 / 
31.8&  56.9 / 51.5&48.2 / 40.1\\
  PromptStyler  /  {\small ICCV-23} \cite{cho2023promptstyler} & 38.7 / 34.0& 45.1 / 34.3& 39.6 / 32.0& 47.0 / 26.8&42.1
 / 25.4& 42.0 / 24.1& 41.8 / 34.4 &42.3 / 30.1\\
        SPG  /  {\small ECCV-24} \cite{bai2024soft} &54.7 / 39.9& 49.8 / 34.9&33.4 / 26.6&47.1 / 41.4&61.3 / 51.1&43.8 / 48.1&58.3 / 52.3&49.8 / 42.0 \\
    MMRL /  {\small CVPR-25} \cite{guo2025mmrl}&44.4 / 37.9&36.9 / 32.4&32.7 / 34.0&44.3 / 43.7&53.7 / 60.8&49.5 / 42.5&59.5 / 66.5 &45.9 / 45.4\\
    LDC /  {\small CVPR-25} \cite{li2025logits}&41.5 / 39.1&52.8 / 39.3&46.7 / 40.0&42.3 / 32.9&50.9 / 33.1&57.8 / 39.7&54.7 / 39.2&49.5 / 37.6 \\
                 \hline
    OmniVaT /  {\small Ours}&67.9 / 66.1&51.8 / 34.9&47.3 / 40.2&61.6 / 64.5&62.9 / 62.6&56.3 / 58.9&60.7 / 68.1&58.4 / 56.5\\
    \textcolor{red}{$\Delta$ LDC}&\textcolor{red}{26.4} / \textcolor{red}{27.0}&\textcolor{red}{-1.0} / \textcolor{red}{-4.4}&\textcolor{red}{0.6} / \textcolor{red}{0.2}&\textcolor{red}{19.3} / \textcolor{red}{31.6}&\textcolor{red}{12.0} / \textcolor{red}{29.5}&\textcolor{red}{-1.5} / \textcolor{red}{19.2}&\textcolor{red}{6.0} / \textcolor{red}{28.9}&\textcolor{red}{8.9} / \textcolor{red}{18.9}\\

      \hline
             
             \hline
  \end{tabular}
  
}
        \label{T-OFReal}

\end{table*}

\begin{table*}[t]
  \centering
  \caption{Comparison with state-of-the-art methods on the OF 2.0 A $\rightarrow$ X task. “OF 2.0 A $\rightarrow$ X”  represents the evaluation of the other unseen target domains, excluding the seen source OF 2.0 A domain. \textcolor{red}{$\Delta$ LDC}  denotes the improvements of the proposed OmniVaT over LDC. The metrics are accuracy (ACC, \%) and Macro-F1 (F1*, \%).  }
   \resizebox{1.0\textwidth}{!}{
  \begin{tabular}{rccccccc}
   \hline
   
   \hline
  \multirow{2}{*}{Methods  / {\small Venue}} &{\makecell[c]{$\rightarrow$ TAG}} & {\makecell[c]{$\rightarrow$ OF Real}} &{\makecell[c]{$\rightarrow$ TQ-DIGIT}}&{\makecell[c]{ $\rightarrow$ TQ-GelMini}} & {\makecell[c]{$\rightarrow$ TQ-DuraGel}}&{\makecell[c]{ $\rightarrow$ TQ-T3D}} &{Average}\\ 
  \cline{2-8}
  ~  & ACC / F1* & ACC / F1* & ACC / F1*& ACC / F1*& ACC / F1* & ACC / F1*& ACC / F1* \\ 
        \hline

   ~ &    \multicolumn{7}{c}{ \textit{ViT-B / 16 \cite{dosovitskiy2020image} with pre-trained weights  from CLIP \cite{radford2021learning}}} \\
            \hline

  CLIP  /  {\small ICML-21} \cite{radford2021learning} &  51.4 / 49.8 &49.0 / 48.8&37.3 / 32.6 & 47.5 / 31.8&44.4 / 
31.8&  56.9 / 51.5&47.8 / 41.1\\
  PromptStyler  /  {\small ICCV-23} \cite{cho2023promptstyler} &  49.4 / 43.0&48.6 / 37.0& 40.4 / 29.4& 55.0 / 50.7& 51.8 / 47.5& 48.1 / 39.3 &48.9 / 41.2\\
        SPG  /  {\small ECCV-24} \cite{bai2024soft} &41.3 / 36.0& 40.7 / 36.4 &57.8 / 46.8&60.4 / 53.3&52.1 / 42.6&42.3 / 41.7&49.1 / 42.8\\
    MMRL /  {\small CVPR-25} \cite{guo2025mmrl}&39.9 / 27.6& 40.9 / 23.6&43.5 / 34.1  &42.8 / 39.2&46.4 / 25.4&43.8 / 30.5 &42.9 / 30.1\\
    LDC /  {\small CVPR-25} \cite{li2025logits}&49.2 / 42.5&50.2 / 40.9&51.9 / 47.1&61.4 / 54.3&54.5 / 44.5&46.1 / 33.8&52.2 / 43.9\\
                 \hline

    OmniVaT /  {\small Ours}&51.0 / 45.5&51.8 / 43.0&64.8 / 66.5&65.1 / 68.2&62.7 / 65.6&56.4 / 60.7&58.6 / 58.2\\
    \textcolor{red}{$\Delta$ LDC}&\textcolor{red}{1.8} / \textcolor{red}{3.0}&\textcolor{red}{1.6} / \textcolor{red}{2.1}&\textcolor{red}{12.9} / \textcolor{red}{19.4}&\textcolor{red}{3.7} / \textcolor{red}{13.9}&\textcolor{red}{8.2} / \textcolor{red}{21.1}&\textcolor{red}{10.3} / \textcolor{red}{26.9}&\textcolor{red}{6.4} / \textcolor{red}{14.3}\\

      \hline
             
             \hline
  \end{tabular}
  
}
        \label{T-OF20A}

\end{table*}

\begin{table*}[t]
  \centering
  \caption{Comparison with state-of-the-art methods on the OF 2.0 B $\rightarrow$ X task. “OF 2.0 B $\rightarrow$ X”  represents the evaluation of the other unseen target domains, excluding the seen source OF 2.0 B domain.  \textcolor{red}{$\Delta$ LDC}  denotes the improvements of the proposed OmniVaT over LDC. The metrics are accuracy (ACC, \%) and Macro-F1 (F1*, \%).  }
   \resizebox{1.0\textwidth}{!}{
  \begin{tabular}{rccccccc}
   \hline
   
   \hline
  \multirow{2}{*}{Methods  / {\small Venue}} &{\makecell[c]{$\rightarrow$ TAG}} & {\makecell[c]{$\rightarrow$ OF Real}} &{\makecell[c]{$\rightarrow$ TQ-DIGIT}}&{\makecell[c]{ $\rightarrow$ TQ-GelMini}} & {\makecell[c]{$\rightarrow$ TQ-DuraGel}}&{\makecell[c]{ $\rightarrow$ TQ-T3D}} &{Average}\\ 
  \cline{2-8}
  ~  & ACC / F1* & ACC / F1* & ACC / F1*& ACC / F1*& ACC / F1* & ACC / F1*& ACC / F1* \\ 
        \hline

   ~ &    \multicolumn{7}{c}{ \textit{ViT-B / 16 \cite{dosovitskiy2020image} with pre-trained weights  from CLIP \cite{radford2021learning}}} \\
            \hline

  CLIP  /  {\small ICML-21} \cite{radford2021learning} &  51.4 / 49.8 &49.0 / 48.8&37.3 / 32.6 & 47.5 / 31.8&44.4 / 
31.8&  56.9 / 51.5&47.8 / 41.1\\

  PromptStyler  /  {\small ICCV-23} \cite{cho2023promptstyler} &  33.8 / 27.8&35.2 / 20.9& 34.7 / 23.8& 32.8 / 17.3& 30.8 / 13.5& 60.5 / 56.7&38.0 / 26.7\\
        SPG  /  {\small ECCV-24} \cite{bai2024soft} &56.8 / 53.0&49.0 / 42.2&53.6 / 45.6& 58.2 / 46.0&42.9 / 37.1&35.9 / 27.9&49.4 / 42.0 \\
    MMRL /  {\small CVPR-25} \cite{guo2025mmrl}&49.3 / 44.1& 45.5 / 36.6&47.3 / 40.2&47.7 / 46.9&23.6 / 21.8&40.3 / 33.8 & 42.3 / 37.2\\
    LDC /  {\small CVPR-25} \cite{li2025logits}&49.2 / 47.9&49.7 / 50.6&51.3 / 41.0&52.4 / 44.2&57.4 / 43.6&53.0 / 35.6&52.2 / 
43.8\\
             \hline
    OmniVaT /  {\small Ours}&65.0 / 62.6&57.3 / 52.4&59.5 / 60.0&64.1 / 65.1&59.2 / 54.4&62.4 / 67.2&61.2 / 60.3\\
    \textcolor{red}{$\Delta$ LDC}&\textcolor{red}{15.8} / \textcolor{red}{14.7}&\textcolor{red}{7.6} / \textcolor{red}{1.8}&\textcolor{red}{8.2} / \textcolor{red}{19.0}&\textcolor{red}{11.7} / \textcolor{red}{20.9}&\textcolor{red}{1.8} / \textcolor{red}{10.8}&\textcolor{red}{9.4} / \textcolor{red}{31.6}&\textcolor{red}{9.0} / \textcolor{red}{16.5}\\

      \hline
             
             \hline
  \end{tabular}
  
}
        \label{T-OF20B}

\end{table*}

\begin{table*}[t]
  \centering
  \caption{Comparison with state-of-the-art methods on the TQ-DIGIT $\rightarrow$ X task. “TQ-DIGIT $\rightarrow$ X”  represents the evaluation of the other unseen target domains, excluding the seen source TQ-DIGIT domain.  \textcolor{red}{$\Delta$ LDC}  denotes the improvements of the proposed OmniVaT over LDC. The metrics are accuracy (ACC, \%) and Macro-F1 (F1*, \%).  }
   \resizebox{1.0\textwidth}{!}{
  \begin{tabular}{rcccccccc}
   \hline
   
   \hline
  \multirow{2}{*}{Methods  / {\small Venue}} &{\makecell[c]{$\rightarrow$ TAG}} &{\makecell[c]{$\rightarrow$ OF Real}} & {\makecell[c]{$\rightarrow$ OF 2.0 A}} & {\makecell[c]{ $\rightarrow$ OF 2.0 B}}&{\makecell[c]{ $\rightarrow$ TQ-GelMini}} & {\makecell[c]{$\rightarrow$ TQ-DuraGel}}&{\makecell[c]{ $\rightarrow$ TQ-T3D}} &{Average}\\ 
  \cline{2-9}
  ~  & ACC / F1* & ACC / F1* & ACC / F1*& ACC / F1*& ACC / F1*& ACC / F1* & ACC / F1*& ACC / F1* \\ 
        \hline

   ~ &    \multicolumn{8}{c}{ \textit{ViT-B / 16 \cite{dosovitskiy2020image} with pre-trained weights  from CLIP \cite{radford2021learning}}} \\
            \hline

  CLIP  /  {\small ICML-21} \cite{radford2021learning} & 53.0 / 40.4&46.7 / 40.6&45.8 / 26.1& 41.2 / 33.7 & 57.3 / 34.1&37.3 / 
30.7&  56.9 / 51.5&48.3 / 36.7\\
  PromptStyler  /  {\small ICCV-23} \cite{cho2023promptstyler} & 52.4 / 47.7& 61.3 / 50.5& 52.0 / 36.8& 45.4 / 38.6& 58.9 / 53.3& 58.6 / 46.7& 51.8 / 56.8 &54.3 / 47.2\\
        SPG  /  {\small ECCV-24} \cite{bai2024soft} &53.4 / 50.8&56.4 / 51.1&49.9 / 37.7&42.4 / 35.7&60.5 / 57.9&55.1 / 45.0&54.1 / 50.0 &53.1 / 46.9\\
    MMRL /  {\small CVPR-25} \cite{guo2025mmrl}&28.4 / 17.3&31.9 / 12.8&27.8 / 10.9&26.8 / 13.5&50.3 / 26.4&46.7 / 23.5&39.9 / 21.1 &36.0 / 17.9\\
    LDC /  {\small CVPR-25} \cite{li2025logits}&53.4 / 51.7&56.1 / 49.6&54.9 / 25.3&44.7 / 33.3&54.9 / 38.6&55.4
/ 41.9&57.0 / 32.1&53.8 / 38.9\\
             \hline
    OmniVaT /  {\small Ours}&55.7 / 52.2&61.8 / 51.7&57.8 / 36.6&44.3 / 35.2&57.8 / 52.8&58.2 / 63.5&58.9 / 59.0&56.4 / 50.1\\
    \textcolor{red}{$\Delta$ LDC}&\textcolor{red}{2.3} / \textcolor{red}{0.5}&\textcolor{red}{5.7} / \textcolor{red}{2.1}&\textcolor{red}{2.9} / \textcolor{red}{11.3}&\textcolor{red}{-0.4} / \textcolor{red}{1.9}&\textcolor{red}{2.9} / \textcolor{red}{14.2}&\textcolor{red}{2.8} / \textcolor{red}{21.6}&\textcolor{red}{1.9} / \textcolor{red}{26.9}&\textcolor{red}{2.6} / \textcolor{red}{11.2}\\

      \hline
             
             \hline
  \end{tabular}
  
}
        \label{T-TQDI}

\end{table*}

\begin{table*}[t]
  \centering
  \caption{Comparison with state-of-the-art methods on the TQ-GelMini $\rightarrow$ X task. “TQ-GelMini $\rightarrow$ X”  represents the evaluation of the other unseen target domains, excluding the seen source TQ-GelMini domain.  \textcolor{red}{$\Delta$ LDC}  denotes the improvements of the proposed OmniVaT over LDC. The metrics are accuracy (ACC, \%) and Macro-F1 (F1*, \%).  }
   \resizebox{1.0\textwidth}{!}{
  \begin{tabular}{rcccccccc}
   \hline
   
   \hline
  \multirow{2}{*}{Methods  / {\small Venue}} &{\makecell[c]{ $\rightarrow$ TAG}} &{\makecell[c]{$\rightarrow$ OF Real}} & {\makecell[c]{$\rightarrow$ OF 2.0 A}} & {\makecell[c]{ $\rightarrow$ OF 2.0 B}}&{\makecell[c]{$\rightarrow$ TQ-DIGIT}}& {\makecell[c]{$\rightarrow$ TQ-DuraGel}}&{\makecell[c]{ $\rightarrow$ TQ-T3D}} &{Average}\\ 
  \cline{2-9}
  ~  & ACC / F1* & ACC / F1* & ACC / F1*& ACC / F1*& ACC / F1*& ACC / F1* & ACC / F1*& ACC / F1* \\ 
        \hline
	 	 
   ~ &    \multicolumn{8}{c}{ \textit{ViT-B / 16 \cite{dosovitskiy2020image} with pre-trained weights  from CLIP \cite{radford2021learning}}} \\
            \hline

  CLIP  /  {\small ICML-21} \cite{radford2021learning} & 53.0 / 40.4&46.7 / 40.6&45.8 / 26.1& 41.2 / 33.7 & 44.4 / 39.1&37.3 / 
30.7& 56.9 / 51.5& 46.5 / 37.4\\

  PromptStyler  /  {\small ICCV-23} \cite{cho2023promptstyler} & 35.8 / 33.2& 55.7 / 45.2& 50.7 / 36.2& 44.3 / 37.1& 43.3 / 46.2& 47.5 / 35.0& 52.0 / 45.6&47.0 / 39.8\\
        SPG  /  {\small ECCV-24} \cite{bai2024soft} &53.7 / 50.0& 56.4 / 48.7& 51.2 / 38.0&40.5 / 34.9&60.1 / 57.9&54.5 / 51.5&60.4 / 59.5&53.8 / 48.6\\
    MMRL /  {\small CVPR-25} \cite{guo2025mmrl}&32.1 / 22.7&32.1 / 12.7 & 27.8 / 11.0&24.0 / 12.3&11.6 / 8.4& 30.7 / 19.7&33.0 / 19.4 &27.3 / 15.2\\
    LDC /  {\small CVPR-25} \cite{li2025logits}&52.9 / 48.5&44.7 / 28.2&47.1 / 32.8&36.4 / 28.0&43.1 / 33.3&41.2 / 33.9&47.6
/ 42.2&44.7 / 35.3\\
             \hline

    OmniVaT /  {\small Ours}&58.0 / 60.0&61.1 / 49.8&53.6 / 39.8&45.1 / 33.9&64.9 / 69.6&56.3 / 60.5&62.0 / 61.0&57.3 / 53.5\\
    \textcolor{red}{$\Delta$ LDC}& \textcolor{red}{5.1} / \textcolor{red}{11.5}&\textcolor{red}{16.4} / \textcolor{red}{21.6}&\textcolor{red}{6.5} / \textcolor{red}{7.0}&\textcolor{red}{8.7} / \textcolor{red}{5.9}&\textcolor{red}{21.8} / \textcolor{red}{36.3}&\textcolor{red}{15.1} / \textcolor{red}{26.6}&\textcolor{red}{14.4} / \textcolor{red}{18.8}&\textcolor{red}{12.6} / \textcolor{red}{18.2}\\

      \hline
             
             \hline
  \end{tabular}
  
}
        \label{T-TQGEL}

\end{table*}

\begin{table*}[t]
  \centering
  \caption{Comparison with state-of-the-art methods on the TQ-DuraGel $\rightarrow$ X task. “TQ-DuraGel $\rightarrow$ X”  represents the evaluation of the other unseen target domains, excluding the seen source TQ-DuraGel domain.  \textcolor{red}{$\Delta$ LDC}  denotes the improvements of the proposed OmniVaT over LDC. The metrics are accuracy (ACC, \%) and Macro-F1 (F1*, \%).  }
   \resizebox{1.0\textwidth}{!}{
  \begin{tabular}{rcccccccc}
   \hline
   
   \hline
  \multirow{2}{*}{Methods  / {\small Venue}}  & {\makecell[c]{$\rightarrow$ TAG}}&{\makecell[c]{$\rightarrow$ OF Real}} & {\makecell[c]{$\rightarrow$ OF 2.0 A}} & {\makecell[c]{ $\rightarrow$ OF 2.0 B}}&{\makecell[c]{$\rightarrow$ TQ-DIGIT}}&{\makecell[c]{ $\rightarrow$ TQ-GelMini}}&{\makecell[c]{ $\rightarrow$ TQ-T3D}} &{Average}\\ 
  \cline{2-9}
  ~  & ACC / F1* & ACC / F1* & ACC / F1*& ACC / F1*& ACC / F1*& ACC / F1* & ACC / F1*& ACC / F1* \\ 
        \hline

   ~ &    \multicolumn{8}{c}{ \textit{ViT-B / 16 \cite{dosovitskiy2020image} with pre-trained weights  from CLIP \cite{radford2021learning}}} \\
            \hline

CLIP  /  {\small ICML-21} \cite{radford2021learning} & 53.0 / 40.4&46.7 / 40.6&45.8 / 26.1& 41.2 / 33.7 & 44.4 / 39.1&57.3 /
34.1  & 37.3 / 
30.7&46.5 / 35.0\\

  PromptStyler  /  {\small ICCV-23} \cite{cho2023promptstyler} & 46.9 / 46.0& 58.4 / 47.7& 47.9 / 36.6& 41.8 / 38.2& 54.3 / 59.1& 53.4 / 47.5& 59.6 / 61.9 &51.8 / 48.1\\
        SPG  /  {\small ECCV-24} \cite{bai2024soft} &54.0 / 53.1& 50.6 / 43.2&38.8 / 34.9&40.7 / 34.6&65.2 / 63.8&56.2 / 44.1&55.5 / 50.2&51.6 / 46.3\\
    MMRL /  {\small CVPR-25} \cite{guo2025mmrl}&25.0 / 10.6&31.9 / 12.1&27.8 / 11.0&25.0 / 10.0&29.0 / 12.0&36.6 / 26.2&32.1 / 19.0 & 29.6 / 14.4\\
    LDC /  {\small CVPR-25} \cite{li2025logits}&53.3 / 51.7&38.4 / 29.8&47.3 / 26.1&43.8 / 25.2&52.6 / 48.4&54.4 / 50.5&54.7 / 36.6&49.2 / 38.3\\
                 \hline
    OmniVaT /  {\small Ours}&57.2 / 57.0&58.3 / 49.8&47.8 / 39.8&49.3 / 36.9&64.1 / 69.6&56.0 / 49.3&61.2 / 60.0&56.3 / 51.8\\
    \textcolor{red}{$\Delta$ LDC}&\textcolor{red}{3.9} / \textcolor{red}{5.3}&\textcolor{red}{19.9} / \textcolor{red}{20.0}&\textcolor{red}{0.5} / \textcolor{red}{13.7}& \textcolor{red}{5.5} / \textcolor{red}{11.7}&\textcolor{red}{11.5} / \textcolor{red}{21.2}&\textcolor{red}{1.6} / \textcolor{red}{-1.2}&\textcolor{red}{6.5} / \textcolor{red}{23.4}&\textcolor{red}{7.1} / \textcolor{red}{13.5}\\

      \hline
             
             \hline
  \end{tabular}
  
}
        \label{T-TQDUR}

\end{table*}

\begin{table*}[t]
  \centering
  \caption{Comparison with state-of-the-art methods on the TQ-T3D $\rightarrow$ X task. “TQ-T3D $\rightarrow$ X”  represents the evaluation of the other unseen target domains, excluding the seen source TQ-T3D domain. \textcolor{red}{$\Delta$ LDC}  denotes the improvements of the proposed OmniVaT over LDC. The metrics are accuracy (ACC, \%) and Macro-F1 (F1*, \%).}
   \resizebox{1.0\textwidth}{!}{
  \begin{tabular}{rcccccccc}
   \hline
   
   \hline
  \multirow{2}{*}{Methods  / {\small Venue}} &{\makecell[c]{ $\rightarrow$ TAG}} &{\makecell[c]{$\rightarrow$ OF Real}} & {\makecell[c]{$\rightarrow$ OF 2.0 A}} & {\makecell[c]{ $\rightarrow$ OF 2.0 B}}&{\makecell[c]{$\rightarrow$ TQ-DIGIT}}&{\makecell[c]{ $\rightarrow$ TQ-GelMini}} & {\makecell[c]{$\rightarrow$ TQ-DuraGel}}&{Average}\\ 
  \cline{2-9}
  ~  & ACC / F1* & ACC / F1* & ACC / F1*& ACC / F1*& ACC / F1*& ACC / F1* & ACC / F1*& ACC / F1* \\ 
        \hline

   ~ &    \multicolumn{8}{c}{ \textit{ViT-B / 16 \cite{dosovitskiy2020image} with pre-trained weights  from CLIP \cite{radford2021learning}}} \\
            \hline

  CLIP  /  {\small ICML-21} \cite{radford2021learning} & 53.0 / 40.4&46.7 / 40.6&45.8 / 26.1& 41.2 / 33.7& 24.2 / 20.4 & 29.5 / 27.4 & 57.5 /  27.3&42.6 / 30.8 \\
  PromptStyler  /  {\small ICCV-23} \cite{cho2023promptstyler} & 52.9 / 48.8& 50.7 / 37.1& 43.3 / 21.4& 44.7 / 26.1& 51.0 / 26.0& 50.8 / 22.0& 52.6 / 35.2 &49.4 / 30.9\\
        SPG  /  {\small ECCV-24} \cite{bai2024soft} &54.3 / 53.8&46.4 / 42.2&45.4 / 27.8&36.2 / 32.9 &59.4 / 48.6&56.8 / 50.1&54.4 / 50.9&50.4 / 43.8 \\
    MMRL /  {\small CVPR-25} \cite{guo2025mmrl}&38.3 / 33.6&56.6 / 46.4&47.5 / 31.1&40.9 / 25.9&53.7 / 37.5&53.9 / 42.9&50.9 / 26.2 &48.8 / 34.8\\
    LDC /  {\small CVPR-25} \cite{li2025logits}&56.1 / 43.2&55.4 / 36.8&46.5 / 21.3&39.2 / 22.5&54.9 / 35.6&58.2 / 42.3&60.2 / 42.4&52.9 / 34.9\\
                 \hline
    OmniVaT /  {\small Ours}&56.5 / 59.1&58.6 / 49.3&47.1 / 30.2&42.2 / 38.1&57.9 / 50.9&59.2 / 54.7&57.1 / 53.3&54.1 / 47.9\\
    \textcolor{red}{$\Delta$ LDC}&\textcolor{red}{0.4} / \textcolor{red}{15.9}&\textcolor{red}{3.2} / \textcolor{red}{12.5}&\textcolor{red}{0.6} / \textcolor{red}{8.9}&\textcolor{red}{3.0} / \textcolor{red}{15.6}&\textcolor{red}{3.0} / \textcolor{red}{15.3}&\textcolor{red}{1.0} / \textcolor{red}{12.4}&\textcolor{red}{-3.1} / \textcolor{red}{10.9}&\textcolor{red}{1.2} / \textcolor{red}{13.0}\\
      \hline
      
             \hline
  \end{tabular}
}
        \label{T-TQ3D}
\end{table*}

\end{document}